\documentclass{article}

\usepackage[preprint]{neurips_2026}

\usepackage[utf8]{inputenc} 
\usepackage[T1]{fontenc}    
\usepackage{hyperref}       
\usepackage{url}            
\usepackage{booktabs}       
\usepackage{amsfonts}       
\usepackage{nicefrac}       
\usepackage{microtype}      
\usepackage{xcolor}         

\usepackage{amsmath}
\usepackage{amssymb}
\usepackage{mathtools}
\usepackage{amsthm}
\usepackage{subcaption}
\usepackage{todonotes}
\usepackage{wrapfig}
\usepackage{enumitem}
\usepackage{pifont}
\usepackage[normalem]{ulem}

\renewcommand{\cite}[1]{\citep{#1}}
\renewcommand{\eqref}[1]{(\ref{#1})}

\title{It Takes Two: Your GRPO Is Secretly DPO}

\author{%
  \small
  \textbf{Yihong Wu}\textsuperscript{1}\thanks{Equal contribution.}\thanks{Correspondence: \texttt{yihong.wu@umontreal.edu}, \texttt{liheng.ma@mail.mcgill.ca}.},\,
  \textbf{Liheng Ma}\textsuperscript{2,3}\footnotemark[1],\,
  \textbf{Lei Ding}\textsuperscript{4},\,
  \textbf{Muzhi Li}\textsuperscript{5},\,
  \textbf{Xinyu Wang}\textsuperscript{2},\,
  \textbf{Kejia Chen}\textsuperscript{6},\,
  \textbf{Zhan Su}\textsuperscript{1}\\
  \textbf{Chenyang Huang}\textsuperscript{7,8},\,
  \textbf{Zhanguang Zhang}\textsuperscript{9},\,
  \textbf{Derek Li}\textsuperscript{9},\,
  \textbf{Yingxue Zhang}\textsuperscript{9},\,
  \textbf{Jian-Yun Nie}\textsuperscript{1},\,
  \textbf{Mark Coates}\textsuperscript{2,3}\\[2pt]
  \normalfont\small
  \textsuperscript{1}UdeM\quad
  \textsuperscript{2}McGill\quad
  \textsuperscript{3}Mila\quad
  \textsuperscript{4}UManitoba\quad
  \textsuperscript{5}CUHK\quad
  \textsuperscript{6}ZJU\quad
  \textsuperscript{7}UAlberta\quad
  \textsuperscript{8}Amii\quad
  \textsuperscript{9}Huawei Noah's Ark Lab%
}

\theoremstyle{plain}
\newtheorem{theorem}{Theorem}[section]
\newtheorem{proposition}[theorem]{Proposition}
\newtheorem{lemma}[theorem]{Lemma}

\theoremstyle{definition}
\newtheorem{definition}[theorem]{Definition}

\theoremstyle{remark}

\usepackage{amsmath,amsfonts,bm}
\usepackage{bbm}

\def\eqref#1{equation~\ref{#1}}

\def\1{\bm{1}}

\def\vx{{\bm{x}}}
\def\vy{{\bm{y}}}

\DeclareMathAlphabet{\mathsfit}{\encodingdefault}{\sfdefault}{m}{sl}
\SetMathAlphabet{\mathsfit}{bold}{\encodingdefault}{\sfdefault}{bx}{n}

\def\gD{{\mathcal{D}}}

\newcommand{\E}{\mathbb{E}}

\newcommand{\Var}{\mathrm{Var}}

\newcommand{\Cov}{\mathrm{Cov}}

\def\E{{\mathbb{E}}}
\newcommand{\llm}{\pi_\theta}
\newcommand{\llmold}{\pi_{\theta_\text{old}}}

\newcommand{\clip}{\mathcal{C}_\epsilon}

\newcommand{\pg}{\bm{g}}

\usepackage{hyperref}
\usepackage{url}
\usepackage{booktabs} 
\usepackage{siunitx}  
\usepackage{makecell} 
\usepackage{multirow} 

\usepackage{graphicx} 

\usepackage{subcaption} 
\usepackage{dsfont}
\usepackage[table]{xcolor}
\usepackage{enumitem}

\usepackage{caption}

\renewcommand{\eqref}[1]{Eq.~(\ref{#1})}

\usepackage[most]{tcolorbox}
\definecolor{bgpurple}{HTML}{F3F2F8}

\newtcolorbox{takeaway}[2][]{
    enhanced, 
    colframe=black, 
    colback=bgpurple, 
    boxrule=1.2pt, 
    arc=4pt, 
    title=#2,
    coltitle=white, 
    colbacktitle=black, 
    fonttitle=\normalsize, 
    attach boxed title to top left={xshift=15pt, yshift=-\tcboxedtitleheight/2},
    boxed title style={
        size=small, 
        boxrule=0pt, 
        arc=2pt 
    },
    top=12pt,
    bottom=8pt,
    left=5pt,
    right=5pt,
    #1 
}

\begin{document}

\maketitle

\begin{abstract}
GRPO has emerged as a prominent reinforcement learning algorithm for post-training LLMs.
Unlike critic-based methods, GRPO computes advantages by estimating the \emph{value baselines} from group-level statistics, eliminating the need for a critic network. 
Consequently, the prevailing view emphasizes the necessity of large group sizes, which are assumed to yield more accurate statistical estimates.
In this paper, we propose a different view that the efficacy of GRPO stems from its implicit contrastive objective in the optimization, which helps reduce variance via the control variate method.
This makes GRPO structurally related to preference learning methods such as DPO.
This perspective motivates 2-GRPO, a minimal group-size variant that constructs contrastive signals with only two rollouts.
We provide a rigorous theoretical analysis of 2-GRPO and empirically validate its effectiveness: 2-GRPO retains $97.6\%$ of the performance of 16-GRPO, while requiring only $12.5\%$ of the rollouts and $21\%$ of the training time.
\end{abstract}

\begin{figure}[h]
    \centering
    
    
    \begin{minipage}[c]{0.605\textwidth}
        \centering
        \includegraphics[width=\textwidth]{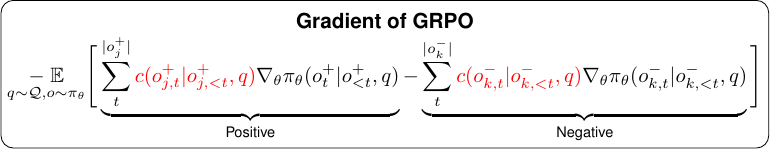}
        
        \vspace{0.2em} 
        
        \includegraphics[width=\textwidth]{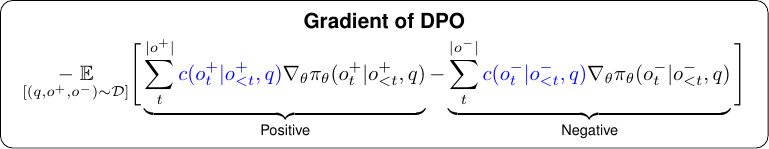}
    \end{minipage}
    \hfill 
    \begin{minipage}[r]{0.38\textwidth}
        \centering
        \includegraphics[width=\textwidth]{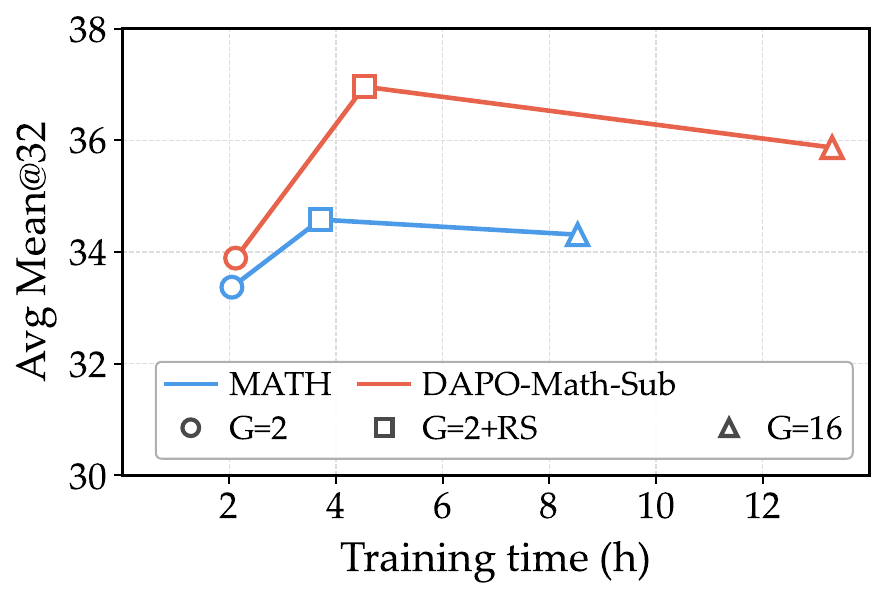}
    \end{minipage}
    
    \vspace{0.4em} 
    
    \includegraphics[width=\textwidth]{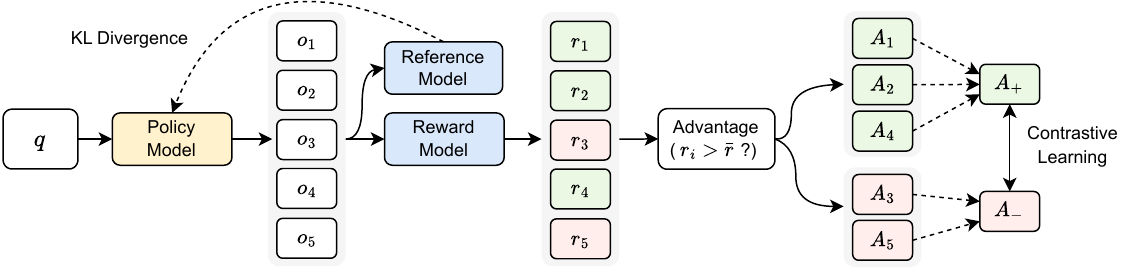}
    
    \caption{\textbf{(Top Left)} Gradients of GRPO and DPO. The two share the same positive--negative structure and differ only in the token-level coefficient. \textbf{(Top Right)} Avg. Mean@$32$ of RL post-trained Qwen-1.5B. 2-GRPO matches the performance of 16-GRPO while substantially reducing training time.
    With resampling, 2-GRPO can even surpass 16-GRPO.
    \textbf{(Bottom)} A conceptual view of GRPO as contrastive (preference) learning: rollouts are split by their advantage relative to the group mean into positive ($A_+$) and negative ($A_-$) sets, which form implicit contrastive pairs.
    }
    \label{fig:main_figure}
\end{figure}

\section{Introduction}
Reinforcement Learning (RL) has emerged as a central paradigm for the post-training of Large Language Models (LLMs).
Two critical functions are aligning model outputs with human intent via RL with Human Feedback (RLHF)~\cite{ouyang2022training} and enhancing reasoning capabilities through RL with Verifiable Rewards (RLVR)~\cite{deepseekai2025deepseekr1incentivizingreasoningcapability}.
Among recent advances, \emph{Group Relative Policy Optimization} (GRPO)~\cite{shao2024deepseekmath} is a prominent critic-free variant of \textit{Proximal Policy Optimization} (PPO)~\cite{schulman2017proximal},
which effectively reduces the variance of gradient estimates by subtracting the estimated value baseline.
Diverging from PPO, which relies on an auxiliary critic network for estimating the value baselines, 
GRPO estimates the advantage function by sampling a group of responses (rollouts) for a single prompt and normalizing their rewards based on the group statistics (mean/standard deviation).
This design eliminates the memory and computational overhead of the value network while maintaining strong performance across various reasoning tasks.

Conventional intuition suggests that GRPO's efficacy is strongly correlated with its group size, grounded in the premise that larger sample sizes yield more accurate advantage estimates and lead to stronger post-trained LLMs.
However, this intuition overlooks the specific construction of the group-relative gradient estimator in GRPO.
First, we demonstrate that GRPO intrinsically functions as contrastive learning~\cite{chopra2005LearningSimilarityMetric} and the \textit{contrastive objective} effectively reduces the variance of the gradient estimates as a control variate method~\cite{johnson2013accelerating}. 
The group of rollouts serves primarily to pair contrastive samples, rather than to estimate the value baselines.
Specifically, the GRPO objective is \textit{de facto} a Monte Carlo estimator to approximate the true contrastive gradients. 
Thus, the choices of group size primarily affects the variance of the Monte Carlo estimator, while the approximation itself remains unbiased. 
Therefore, in contrast to the prevailing value-baseline estimation viewpoint, GRPO with a small group size shall still work properly.
Second, this perspective further reveals its close connection to the well-known \textit{Direct Preference Optimization}~(DPO) algorithm~\cite{rafailov2023direct}, which explicitly introduces the contrastive objective in the offline RLHF setting.
The GRPO is \textit{de facto} doing \textit{direct preference optimization} on online RL settings with the necessary adaptations.

To evidence this hypothesis, 
we propose the minimal two-rollout setting (2-GRPO), 
a configuration previously regarded as inadequate for estimating group statistics~\cite{student1908probable}, 
but well aligned with the contrastive learning interpretation and the DPO objectives.
We provide a thorough theoretical analysis of the properties of 2-GRPO and empirically evaluate its effectiveness and efficiency across a diverse set of models and tasks. 
The theoretical analysis justifies the rationale behind the 2-GRPO designs.
Empirically, 2-GRPO achieves performance comparable to 16-GRPO while substantially reducing training time. 
We further propose a resampling variant, 2-GRPO+RS, which reduces sample discard rate and achieves performance closer to 16-GRPO while being more efficient than 16-GRPO.
These findings support our central hypothesis: GRPO derives its strength primarily from its contrastive formulation, rather than from accurate advantage estimation.
The efficiency of 2-GRPO further highlights the promise of the contrastive policy optimization direction.


\vspace{-0.2em}

\begin{takeaway}{Overview}
\begin{itemize}[leftmargin=*, itemsep=2pt]

\item \textbf{Theoretical Finding:} GRPO reduces variance of policy gradient estimate via a \textit{Contrastive Gradient Optimization}, rather than through reward shaping with value baselines as in PPO. GRPO is \textit{de facto} the online RL version of DPO with corresponding adaptations.

\item \textbf{Evidence:} GRPO with a group size of 2 is invalid from the value estimate perspective but remains valid under the contrastiveness view. 
Empirically, 2-GRPO remains effective, indicating that contrastive learning provides a more principled explanation.

\item \textbf{Implication:} 2-GRPO matches the performance of 16-GRPO with substantially fewer rollouts, indicating that large group sizes are not essential and that advancing GRPO through contrastive learning techniques is a promising direction.

\end{itemize}

\end{takeaway}
\section{Preliminary}

\subsection{Problem Setting and Notation}

Our work focuses on RL-based post-training of LLMs for reasoning capabilities.
Given an input prompt $q \in \mathcal{Q}$, 
the model generates the $i$-th response $o_i = (o_{i,1}, \ldots, o_{i,T})$, where $o_{i,t}$ is the token generated at step $t \in [0, T]$ and $o_{i,<t}$ denotes the sequence of preceding tokens.
A trajectory $\tau = (q,o) \in \mathcal{T}$ is defined as a concatenation of a prompt and its corresponding generated response.
In current RL post-training, the reward function $r: \mathcal{T} \rightarrow \mathbb{R}$ is typically defined at the trajectory level.
The learning objective is to maximize the expected reward over the trajectory space:
\begin{equation}
\label{eq:obj}
     \mathcal{J}(\theta) = \E_{q \sim \mathcal{Q}}\E_{o \sim \llm (\cdot | q)} [r(\tau)]\,,
\end{equation}
where $\pi_\theta$ denotes the policy model, a LLM with parameters $\theta$; and
$\mathcal{Q}$ is the set of prompts, each consisting of a question and necessary instructions.
We mainly focus on the setting of verifiable rewards, where the responses can be verified as correct ($r=1$) or incorrect ($r=0$).

\subsection{The Story of Variance Reduction: VPG, PPO, and GRPO}

As a foundational policy gradient method, \emph{Vanilla Policy Gradient} (VPG) \cite{williams1992simple} optimizes the objective function using the following gradient estimator (where $r_i$ is the reward of $(q, o_i)$):
\begin{equation}
\label{eq:vpo} 
\nabla_\theta \mathcal{J}(\theta) = \underset{\substack{q \sim \mathcal{Q} \\ o_i \sim \pi_{\theta} } }{\E}
\left[ r_i \sum_{t=0}^{|o_i|} \nabla_\theta \log\llm (o_{i,t} | o_{i,<t}, q) \right] \;.
\end{equation}

Although effective, VPG usually suffers from high variance of gradient estimates and training instability.  
Therefore, subsequent works~\cite{schulman2015trust, schulman2017proximal} utilize advantage estimates~\cite{baird1993advantage} to reduce the variance of the policy gradient estimator: $A_{i,t} = r_i - b(q)$,
where $A_{i,t}$ is token-level advantage and $b(q)$ is the value baseline function~(See Appx.~\ref{appx:var_reduction} for more details).
An auxiliary LLM is employed as a critic to estimate this value baseline, such as in \emph{Proximal Policy Optimization} (PPO) \citep{schulman2017proximal}:
\begin{equation}
\label{eq:ppo}
\mathcal{J}(\theta)
=
\underset{\substack{q \sim \mathcal{Q} \\ o_i \sim \pi_{\theta_{\text{old}}}}}{\mathbb{E}}
\frac{1}{|o_i|}\sum_{t=1}^{|o_i|}
\min \left\{
A_{i,t} \rho_{i,t},  A_{i,t}\mathcal{C}_\epsilon(\rho_{i,t})
\right\} \, , \,
\rho_{i,t}=\frac{\pi_{\theta}\!\left(o_{i,t}\mid o_{i,<t}, q\right)} 
{\pi_{\theta_{\text{old}}}\!\left(o_{i,t}\mid o_{i,<t}, q\right)}
\, ,
\end{equation}
where $\pi_{\theta_{\text{old}}}$ is the policy used to generate trajectories, while $\pi_\theta$ denotes the current policy being optimized.
The $\rho_{i,t}$ term is the introduced importance sampling technique for online (near-)on-policy RL while
$\mathcal{C}_\epsilon(x)$ denotes the clipping function within the interval $[1-\epsilon, 1+\epsilon]$.


To eliminate the substantial computational overhead and memory demands of the critic network, several studies~\cite{li2024remax, ahmadian2024back, shao2024deepseekmath} propose estimating the baseline without a critic network. 
Specifically,
\emph{GRPO} estimates the advantage using the reward statistics from a group of generated responses:
\begin{equation}
    A_{i,t} = \frac{r_i - \text{mean}(\mathbf{r})}{\text{std}(\mathbf{r}) + \epsilon} \, ,
    \label{eq:grpo_adv}
\end{equation}
where $r_i$ is the reward for response $o_i$ given query $q$, and $\mathbf{r}$ denotes the vector of rewards for $G$ sampled responses associated with $q$. 
Therefore, it is generally believed that GRPO requires a sufficiently large group size to obtain accurate group-level statistics for advantage estimation.~\footnote{Due to limited space, the comprehensive related work is provided in Appx.~\ref{appx:related_work}.}

\section{A Tale of Two Algorithms: GRPO and DPO}
\label{sec:rewriting_grpo}

At first glance, the objectives of GRPO and DPO appear distinct on different RL settings.
We show that
they are the twin objects of the contrastive RL objective under the online/offline RL setting,
which can be seen from 
the gradient forms of GRPO and DPO.
This finding provides a new theoretical analysis (Sec.~\ref{sec:ctr_rl}) and motivates a more efficient yet effective algorithm (Sec.~\ref{sec:2-grpo}).

\subsection{Contrastive Objective for Sequences}
\label{sec:contrastive_seq}

\emph{Contrastive Learning}~\cite{chopra2005LearningSimilarityMetric} has been a powerful learning paradigm in (self-)supervised learning,
ranging from $1$-vs-$1$ (one positive and one negative) objectives~\cite{rendle2009bpr} to $1$-vs-$N$~\cite{oord2018representation} and $N$-vs-$M$ variants~\cite{frosst2019analyzing}. 
We first formalize the contrastive loss objective for sequences for further analysis.

\begin{definition}[Contrastive Loss for Sequences]
\label{def:cl}
Let $\pi_\theta$ be a probabilistic model and $\mathcal{D}$ be a data distribution. Consider an anchor sequence $\mathbf{x} \sim \mathcal{D}$, and let $\mathcal{D}^+(\cdot \mid \mathbf{x})$ and $\mathcal{D}^-(\cdot \mid \mathbf{x})$ denote the conditional distributions for positive and negative samples, respectively.
Let $y_{t}$ denote the $t$-th token of sequence $\vy$.
A differentiable loss function $\mathcal{L}$ is \emph{contrastive} if its gradient holds the following form:
\begin{equation}
\begin{aligned}
\label{eq:cl_loss_gradient} 
\nabla_\theta \mathcal{L} &= -\underset{\vx \sim \mathcal{D}}{\E}
\Bigg[
\underset{\vy^+ \sim \mathcal{D}^+}{\E} \sum_{t=1}^{|\vy^+|} c_{t}^+ \nabla_\theta \llm(\vy^+_{t} | \vy^+_{<t}, \vx)  
\quad
-\underset{\vy^- \sim \mathcal{D}^-}{\E} \sum_{t=1}^{|\vy^-|} c_{t}^- \nabla_\theta \llm\left(\vy^-_{t} | \vy^-_{<t}, \vx \right) 
\Bigg] \, ,
\end{aligned}
\end{equation}
where $c_{t}^+$ and $c_{t}^-$ are token-level coefficients depending on specific algorithm design.
\end{definition}
We adopt token-level coefficients for generality, as sequence-level coefficients can be recovered as a special case.  
Furthermore, the number of \textit{positive} ($N$) and \textit{negative} ($M$) samples of each data point may vary depending on the specific designs, serving as a \textit{Monte Carlo estimator} to approximate the true gradient in~\eqref{eq:cl_loss_gradient}.

\paragraph{DPO is a 1-vs-1 contrastive learning} \emph{Direct Preference Optimization}~(DPO)~\cite{rafailov2023direct} is a dominant offline RLHF algorithms for LLMs:
\begin{equation}
    \mathcal{L}_{\text{DPO}} = \underset{(q, o^+, o^-) \sim \mathcal{D}_{\text{DPO}}}{-\mathbb{E}} \left[ \log \sigma \left( \beta \log \frac{\pi_\theta (o^+|q)}{\pi_{\text{ref}} (o^+|q)} - \beta \log \frac{\pi_\theta (o^-|q)}{\pi_{\text{ref}} (o^-|q)}
    \right) \right] \; ,
\label{eq:dpo}
\end{equation}
where the preference pair $(q, o^+, o^-) \sim \mathcal{D}_{\text{DPO}}$ are from precollected human-annotated data. $\sigma$ denotes the sigmoid function.
It is easy to show that DPO is a 1-vs-1 contrastive learning.
We provide Lemma~\ref{lemma:dpo_is_cl} and its proof in Appx.~\ref{appx:dpo_is_cl} for reference.

\subsection{GRPO: N-vs-M Contrastive Learning}
We demonstrate that GRPO effectively functions as a dynamic $N$-vs-$M$ contrastive learning, where the group size $G = N + M$ is fixed, but the specific values of $N$ (positive samples) and $M$ (negative samples) are dynamic based on the sampled responses.
Let $G^+_q$ and $G^-_q$ denote the counts of correct and incorrect trajectories, respectively.
The GRPO objective function can  be formulated as:
\begin{equation}
\label{eq:grpo_finite}
\begin{aligned}
&\mathcal{J}_\text{GRPO}(\theta, G) = {\E}_{\left[q \sim \mathcal{Q}; \{o_j^+, o_k^-\}_{j,k}^{G} \sim \llmold(\cdot|q)\right]} \\&{\sqrt{\widehat{\Var}_G(q)}}   \Bigg[
    \underbrace{\frac{1}{G_q^+}\sum_{j=1}^{G_q^+}  \frac{1}{|o_j^+|}\sum_{t=1}^{|o_j^+|} \clip^+ \left(  \rho_{j,t} \right)}_{\text{positive}}
    - \underbrace{\frac{1}{G_q^-} \sum_{k=1}^{G_q^-} \frac{1}{|o_k^-|} \sum_{t=1}^{|o_k^-|} \clip^- \left( \rho_{k,t} \right)}_{\text{negative}}
    \Bigg] \, ,
\end{aligned}
\end{equation}
where $o_j^+$ and $o_k^-$ denote rollouts with correct and incorrect outcomes, respectively.
Denoting $\hat{p}_{\theta_\text{old},q} = G^+_q/G$, the term $\widehat{\Var}_G(q) = (1-\hat{p}_{\theta_\text{old},q})\hat{p}_{\theta_\text{old},q}$
is the empirical variance of the $G$ sampled trajectories from the true $\text{Bernoulli}(p_{\text{old},q})$ under the RLVR setting.\footnote{In subsequent parts, we omit the subscript $\theta_\text{old}$ of $p$ for brevity.}
For simplicity,
we denote the upper and lower clippings as $\clip^+(x)=\min[x, 1+\epsilon]$ and $\clip^-(x)=\max[x, 1-\epsilon]$, respectively.

The formulation in~\eqref{eq:grpo_finite} provides the foundation for the following proposition, with a proof provided in Appx.~\ref{appx:rewrite_grpo}.
Despite the sophisticated algorithm design of GRPO, this proposition unveils its  \emph{contrastive} nature.
\begin{proposition}
\label{prop:grpo_is_cl}
The maximization of the GRPO objective is equivalent to the minimization of an $N$-vs-$M$ contrastive loss estimator.
\end{proposition}

The derivation of Proposition~\ref{prop:grpo_is_cl} is based on the binary reward assumption to align with the RLVR setting.
However, \textit{GRPO's contrastive nature} extends to continuous rewards inherently (as shown in Figure~\ref{fig:main_figure}).
Due to the space limit, we focus on the properties of GRPO on RLVR.

\subsection{Echoes of Contrastiveness: GRPO and DPO}
\begin{takeaway}{Summary}

GRPO and DPO perform the same \textbf{contrastive policy gradient} -- increasing preferred outputs relative to unpreferred ones -- under different choices of Monte Carlo estimator.
They are instantiated under different RL regimes, leading to different design choices in weighting, aggregation, and regularization.
\end{takeaway}

Based on previous analysis, the differences between GRPO and DPO are merely in the coefficients of contrastive gradient:
\definecolor{tokenW}{HTML}{0B3D91}   
\definecolor{groupW}{HTML}{1B5E20}   
\definecolor{aggR}{HTML}{B71C1C}     
\definecolor{clipRef}{HTML}{4A148C}  
\begin{align}
    \text{GRPO:}\quad &c(o_{i,t}\mid o_{i,<t},q)
    := \frac{\textcolor{tokenW}{\mathbbm{1}^\epsilon_{i,t}}\,
             \textcolor{groupW}{\sqrt{\widehat{\mathrm{Var}}(q)}}}
            {\textcolor{aggR}{|o_i|}\,
             \textcolor{tokenW}{\pi_{\theta_{\text{old}}}(o_{i,t}\mid o_{i,<t},q)}} \,, \\
    \text{DPO:}\quad &c(o_t\mid o_{<t},q)
    := \frac{\beta\,\textcolor{groupW}{\sigma(\hat{r}_\theta(q,o^-) - \hat{r}_\theta(q,o^+))}}
            {\textcolor{tokenW}{\pi_\theta(o)}}\,,
    \quad \hat{r}_\theta = \beta\log\frac{\pi_\theta(y\mid x)}{\textcolor{clipRef}{\pi_{\text{ref}}(y\mid x)}}\,.
\end{align}

In the following, we show that the differences between GRPO and DPO are largely adaptations to their respective learning regimes: GRPO operates online with generated rollouts, whereas DPO operates offline with pre-collected preference data.

\textbf{\textcolor{black}{Group Size}.}
DPO typically learns with fixed $1$-vs-$1$ preference pairs which are collected offline in advance.
By contrast, due to sampling responses online, GRPO needs to handle arbitrary $N$-vs-$M$ positive--negative samples within each group. 
This changes only the Monte Carlo sample size used to estimate the same positive and negative contrastive gradients.

\textbf{\textcolor{aggR}{Token Aggregation}.}
Within a sequence, GRPO averages token-level gradients, whereas DPO sums them. 
This is a design choice rather than a fundamental difference: e.g., SimPO~\cite{meng2024simpo} -- a DPO variant -- uses mean aggregation, while Dr.\ GRPO~\cite{liu2025understanding}-- a GRPO variant -- adopts sum aggregation.

\textbf{\textcolor{tokenW}{Token-Level Weighting} (Importance Sampling vs.\ Log-Likelihood).}
In GRPO, importance-sampling coefficients correct the gradient for samples generated by the old policy, specifically for its near-on-policy online RL setting. It is typically used together with clipping for training stability.
DPO, however, does not require such correction in the offline setting and therefore directly uses the log-likelihood form of $\pi_\theta$.

\textbf{\textcolor{groupW}{Group-Level Weighting}.}
GRPO weights each group by $\sqrt{\widehat{\mathrm{Var}}(q)}$, embodying its design philosophy of attending to more uncertain questions. DPO, in contrast, weights each pair by $\sigma(\hat{r}_\theta(q, o^-) - \hat{r}_\theta(q, o^+))$, assigning higher scores to pairs where the negative sample outscores the positive one.

\textbf{\textcolor{clipRef}{Reference Model}.}
DPO is regularized toward the reference model $\pi_{\text{ref}}$ through an implicit KL term.
Optionally, GRPO can add a separate explicit KL penalty term w.r.t. the reference model.

In conclusion, the differences between GRPO and DPO mainly reflect adaptations to online vs. offline RL settings. 
The core mechanisms remain the same: both estimate a contrastive gradient that increases the likelihood of preferred outputs relative to unpreferred ones.


\section{Why Viewing GRPO From Contrastive Learning?}
\label{sec:ctr_rl}

\begin{takeaway}{Summary}

In this section, we address two key questions:
\begin{itemize}[leftmargin=0pt, itemindent=*, itemsep=1pt, topsep=0pt, parsep=0pt, partopsep=0pt]
\item How contrastive learning reduces the variance of policy gradient estimates in GRPO?
\item Why contrastiveness is more principled than the value-baseline view?
\end{itemize}
\end{takeaway}


\subsection{Variance Reduction via Contrastive Objective}
We demonstrate that this contrastive gradient formulation functions as a control variate method, where the coefficients serve to control the variance of the estimator.
\begin{proposition}\label{prop:control_variate}Let $\pi_\theta$ denote the policy model. Let $o^+ \sim \pi_\theta^+(\cdot|q)$ and $o^- \sim \pi_\theta^-(\cdot|q)$ denote random variables representing a positive sample and a negative sample, respectively.
Let $\bm{g}^+ = \nabla_\theta \log \pi_\theta(o^+ | q)$, $\bm{g}^- = \nabla_\theta \log \pi_\theta (o^- | q)$ and $\rho$ denote the correlation coefficient of $\bm{g}^+$ and $\bm{g}^-$.
If $\Cov(\bm{g}^+ , \bm{g}^-)>0$ and $0 \leq c \leq 2\frac{\Cov(\bm{g}^+ , \bm{g}^-)}{\Var(\bm{g}^-)}$, then $\Var(\bm{g}^+ - c\bm{g}^-) \leq \Var(\bm{g}^+)$.
Specifically, if $c=\frac{\Cov(\bm{g}^+ , \bm{g}^-)}{\Var(\bm{g}^-)}$, then
\begin{equation}
\Var(\bm{g}^+ - c\bm{g}^-) =  \left(1-\rho^2 \right) \Var(\bm{g}^+) \, ,
\end{equation}
where $\text{Var}(\cdot)$ and $\text{Cov}(\cdot, \cdot)$ denotes the corresponding traces of var/cov matrices for gradient vectors.
\end{proposition}


This proposition (proof in Appx.~\ref{proof:control_variate})  shows that, when the coefficient \( c \) lies within an appropriate range, the variance of the gradient estimator can be reduced. This result directly follows the control variate method, a variance reduction technique widely used in Monte Carlo estimation and stochastic gradient optimization~\cite{johnson2013accelerating}.
The reduction of gradient variance stabilizes RL training~\cite{li2024remax}.

A key implication of Proposition~\ref{prop:control_variate} is that the degree of variance reduction depends on the correlation between positive and negative samples. 
In LLM post-training, the positive sample \( o^+ \) and the negative sample \( o^- \) are generated by the same model conditioned on the same prompt \( q \), 
which typically induces a nontrivial correlation between them. (See Appx.~\ref{appx:correlation_pos_neg} for more discussion.)

\subsection{GRPO with Small Group Size: It Should Fail, But Doesn’t}
\label{sec:2-grpo}

While both the value-baseline and contrastive perspectives account for GRPO's variance reduction, they rest on fundamentally different assumptions.
The prevailing value-baseline view holds that GRPO requires a sufficiently large group size to yield reliable group-level statistics; under this view, small-group GRPO should fail due to high-variance estimates~\cite{student1908probable}.
The contrastive perspective, by contrast, treats the positive and negative samples within a group as Monte Carlo estimates of the true positive and negative gradients. Because Monte Carlo estimation is unbiased regardless of sample size, GRPO with small groups should remain effective under stochastic optimization.

To adjudicate between these two views, we introduce 2-GRPO, a variant that uses the minimal group size of $2$.
\textbf{The value-baseline perspective predicts that this setting will fail} (see Appx.~\ref{appx:student} for details).
Empirically, however, 2-GRPO matches the performance of standard GRPO while achieving substantially higher efficiency, supporting the contrastive perspective as a more principled account of GRPO's underlying mechanism.
We describe 2-GRPO concretely in the following section.

\subsection{Introducing 2-GRPO}
\label{sec:2-GRPO}
With a group size of two, the GRPO advantage reduces to a simple contrastive signal: $A^+ = +1$ and $A^- = -1$ when the two rollouts disagree on the reward, and both zeros otherwise. \textbf{This yields an online RL counterpart of Direct Preference Optimization (DPO).}

\newcommand{\cmark}{\ding{51}} 
\newcommand{\xmark}{\ding{55}} 

\begin{takeaway}{Analysis on 2-GRPO}
\textbf{(Appx.~\ref{appx:adv_estimate}) Implicit Reweighting under Stochastic Optimization.}
At first glance, 2-GRPO uses fixed advantages regradless prompt success rates. However, under stochastic optimization, prompts are implicitly reweighted by their probability of forming a contrastive pair.

\textbf{(Appx.~\ref{appx:gradient_estimate}) The Real Driver of Variance Reduction: Mini-Batch Size, Not Group Size.}
A common view attributes GRPO's stability to its large group size. We argue that the key factor is instead the \textit{size of training mini-batch}. Enlarging the group size is one way to increase rollouts, but enlarging the number of prompts per mini-batch achieves the same variance reduction.

\textbf{(Appx.~\ref{appx:breakdown}) Exploration on Hard Questions.}
A natural concern is that small-group GRPO may under-explore difficult prompts. We show that, for a fixed total rollout budget, the probability of 2-GRPO sampling at least one correct answer is no lower than that of 16-GRPO.
\end{takeaway}

\textbf{2-GRPO with Re-Sampling.}
Because of its binary contrastive nature, 2-GRPO discards any group whose two rollouts share the same reward. When the policy is highly accurate on the training set, this wastes a substantial fraction of generated samples and leads to sub-optimal performance.~\footnote{A discussion on discard rate is provided in Appx.~\ref{appx:discard_rate}.}

To address this, we introduce a resampling variant, denoted \textbf{2-GRPO+RS}, which follows the strategy of DAPO~\cite{yu2025dapo}: whenever a group is discarded due to a zero advantage, it is replaced with a fresh group sampled from a new prompt. Resampling adds rollout-stage computation, but the overhead is modest and consistently lifts peak performance.

\textbf{Efficiency Gains.}
The efficiency gains of 2-GRPO arise at two stages: rollout generation and policy optimization.
For a fixed number of prompts, 2-GRPO generates only $12.5\%$ of the rollouts required by 16-GRPO, and optimizes over the same $12.5\%$ fraction during policy updates.
2-GRPO+RS may generate additional rollouts through resampling, but its optimization-stage cost matches standard 2-GRPO. In our experiments, we cap the rollout budget of 2-GRPO+RS at that of 16-GRPO; in practice, it typically uses fewer.

\section{Experiments}
\label{sec:exp}

\textbf{Goal of Experiment.}
Building on the theoretical justification for 2-GRPO, we seek to empirically assess its validity in RLVR.
We anticipate that \textbf{\emph{2-GRPO will achieve a comparable performance as the regular GRPO (16-GRPO), 
and exhibit better efficiency}}—with respect to computational resources and/or wall-clock time.

\textbf{Datasets, Baselines and Hyper-parameters.}
We provide the details of datasets, baselines and hyper-parameter choices in Appx.~\ref{appx:exp_details}.
For training, we adopt the \emph{verl} framework~\cite{sheng2025hybridflow} and utilize the built-in implementation of GRPO~\cite{shao2024deepseekmath} as the baseline algorithm.

\subsection{Math Reasoning}
\label{sec:main_exp}
\definecolor{marketup}{RGB}{0,153,0}    
\definecolor{marketdown}{RGB}{204,0,0}  
\newcommand{\good}[1]{(\textcolor{marketup}{#1})}
\newcommand{\bad}[1]{(\textcolor{marketdown}{#1})}

\begin{table}[t!]
    \centering
    \caption{\small 2-GRPO vs. 16-GRPO: post-trained on MATH/DAPO-Math-Sub and evaluated on five math reasoning benchmarks. $G$ is the group size. $(\cdot)$ shows the gaps to 16-GRPO. All models are post-trained with 10 epochs.}
    \label{tab:2vs16}
    \vspace{0.5em}

    \setlength{\tabcolsep}{6pt}
    \renewcommand{\arraystretch}{1.15}

    \resizebox{\linewidth}{!}{%
    \begin{tabular}{@{}l l r r r r r r r@{}}
        \toprule
        \textbf{Mean@32}$\uparrow$ & \textbf{$G$} & \textbf{Time (h)} $\downarrow$
        & \textbf{MATH-500} & \textbf{AMC 2023} & \textbf{Minerva Math}
        & \textbf{AIME 2025} & \textbf{Olympiad Bench} & \textbf{Avg} \\
        \midrule

        \rowcolor{gray!12}
        \multicolumn{9}{c}{\textit{Post-training on MATH dataset}} \\
        \midrule

        \multicolumn{9}{@{}l}{\textbf{Qwen-1.5B}} \\
        \addlinespace[2pt]
        \quad      & w/o   & --
            & 31.83 & 34.30 & 5.33  & 3.64  & 15.40 & 18.10 \\
        \quad     & 16   & 8.53
            & 70.24 & 51.25 & 16.84 & \textbf{10.10} & 23.11 & 34.31 \\
        \quad Ours    & 2    & 2.05\,\good{-75.96\%}
            & 69.28\,\bad{-0.96}
            & 49.53\,\bad{-1.72}
            & 16.25\,\bad{-0.59}
            & 9.48\,\bad{-0.62}
            & 22.31\,\bad{-0.80}
            & 33.37\,\bad{-0.94} \\
        \quad Ours    & 2+RS & 3.71\,\good{-56.51\%}
            & \textbf{71.36}\,\good{+1.12}
            & \textbf{51.64}\,\good{+0.39}
            & \textbf{18.74}\,\good{+1.90}
            & 7.29\,\bad{-2.81}
            & \textbf{23.85}\,\good{+0.74}
            & \textbf{34.58}\,\good{+0.27} \\
        \addlinespace[2pt]
        \midrule

        \multicolumn{9}{@{}l}{\textbf{Qwen-7B}} \\
        \addlinespace[2pt]
        \quad      & w/o   & --
            & 47.16 & 38.36 & 5.99  & 5.00  & 9.83  & 21.27 \\
        \quad     & 16   & 9.30
            & 75.90 & 61.79 & 22.81 & \textbf{13.23} & 25.99 & 39.94 \\
        \quad Ours    & 2    & 2.43\,\good{-73.87\%}
            & 75.23\,\bad{-0.67}
            & \textbf{64.60}\,\good{+2.81}
            & 23.13\,\good{+0.32}
            & 12.81\,\bad{-0.42}
            & \textbf{26.39}\,\good{+0.40}
            & \textbf{40.43}\,\good{+0.49} \\
        \quad  Ours & 2+RS & 5.78\,\good{-37.85\%}
            & \textbf{76.89}\,\good{+0.99}
            & 61.64\,\bad{-0.15}
            & \textbf{24.90}\,\good{+2.09}
            & 11.67\,\bad{-1.56}
            & 25.39\,\bad{-0.60}
            & 40.10\,\good{+0.16} \\
        \midrule

        \rowcolor{gray!12}
        \multicolumn{9}{c}{\textit{Post-training on DAPO-Math-Sub dataset}} \\
        \midrule

        \multicolumn{9}{@{}l}{\textbf{Qwen-1.5B}} \\
        \addlinespace[2pt]
        \quad      & w/o   & --
            & 31.83 & 34.30 & 5.33  & 3.64  & 15.40 & 18.10 \\
        \quad     & 16   & 13.30
            & 70.66 & 56.56 & 18.00 & 9.58  & 24.56 & 35.87 \\
        \quad Ours    & 2    & 2.12\,\good{-84.06\%}
            & 68.81\,\bad{-1.85}
            & 52.19\,\bad{-4.37}
            & 16.79\,\bad{-1.21}
            & 8.13\,\bad{-1.45}
            & 23.52\,\bad{-1.04}
            & 33.89\,\bad{-1.98} \\
        \quad Ours    & 2+RS & 4.53\,\good{-65.93\%}
            & \textbf{71.64}\,\good{+0.98}
            & \textbf{58.59}\,\good{+2.03}
            & \textbf{20.11}\,\good{+2.11}
            & \textbf{9.79}\,\good{+0.21}
            & \textbf{24.67}\,\good{+0.11}
            & \textbf{36.96}\,\good{+1.09} \\
        \addlinespace[2pt]
        \midrule

        \multicolumn{9}{@{}l}{\textbf{Qwen-7B}} \\
        \addlinespace[2pt]
        \quad      & w/o   & --
            & 47.16 & 38.36 & 5.99  & 5.00  & 9.83  & 21.27 \\
        \quad     & 16   & 17.68
            & 77.35 & \textbf{69.69} & \textbf{24.45} & 14.27 & 28.86 & 42.92 \\
        \quad Ours    & 2    & 3.63\,\good{-79.47\%}
            & \textbf{77.43}\,\good{+0.08}
            & 64.84\,\bad{-4.85}
            & 21.95\,\bad{-2.50}
            & 14.58\,\good{+0.31}
            & \textbf{29.86}\,\good{+1.00}
            & 41.73\,\bad{-1.19} \\
        \quad Ours    & 2+RS    &7.55\,\good{-57.29\%} 
            & 77.14\,\bad{-0.21}
            & 68.91\,\bad{-0.78}
            & 23.94\,\bad{-0.51}
            & \textbf{16.67}\,\good{+2.40}
            & 28.39\,\bad{-0.47}
            & \textbf{43.01}\,\good{+0.09} \\
        \bottomrule
    \end{tabular}}
\end{table}

\begin{figure}[t!]
    \centering
    \includegraphics[width=1\linewidth]{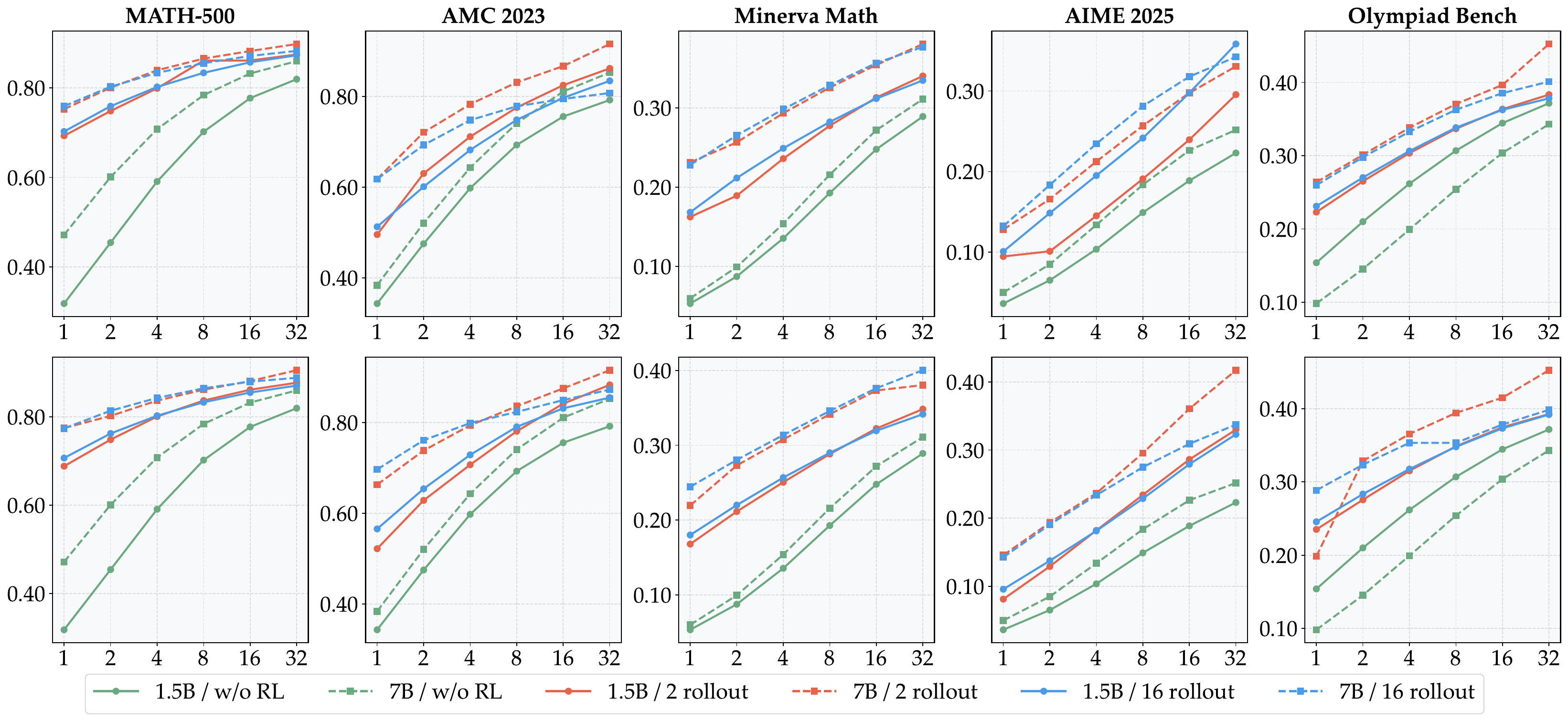}
    \caption{Pass@$K$ ($y$) vs. $K=1,2,\cdots$ ($x$) over five math benchmarks. First/second row for models post-trained on MATH/DAPO-Math-Sub dataset.}
    \label{fig:pass@k}
    \vspace{-0.2cm}
\end{figure}

Following prior studies~\cite{yu2025dapo}, 
we consider mathematical tasks as representative instances of RLVR to verify our hypothesis.
In the main experiment, the models are post-trained with RL techniques on MATH and DAPO-Math-Sub datasets under a fixed budget of 10 training epochs.
The post-trained models are evaluated on five widely-used math reasoning benchmarks.
This is an out-of-distribution evaluation setting, imposing requirements on the generalization ability of the post-trained models.

\textbf{Table~\ref{tab:2vs16} showcases the 
Mean@32 as well as the training time.}
The empirical results show that \uline{2-GRPO achieves 97.6\% of 16-GRPO's average performance while using only 12.5\% of its total rollouts and 21.0\% of its training time}.~\footnote{Appx.~\ref{appx:why_rollouts} discusses the relationship between the total number of rollouts and computational cost.}

These results provide strong corroboration of our theoretical finding that reducing group size preserves performance while substantially improving efficiency.

The resampling variant, \uline{2-GRPO+RS, further improves peak performance, outperforming 16-GRPO on average while using roughly half of its training time}. 
Although it is slower than 2-GRPO, it remains substantially more efficient than 16-GRPO, making it a practical alternative that preserves small-group efficiency while recovering the performance benefits of broader exploration.

\textbf{Figure~\ref{fig:pass@k} shows the Pass@$K$ over various $K$ choices.}
Overall, 2-GRPO achieves Pass@$K$ performance comparable to 16-GRPO across different choices of $K$.
In particular, 2-GRPO even outperforms 16-GRPO on the AMC 2023 and Olympiad Bench. 
On AIME 2025, 2-GRPO performs better when post-trained on DAPO-Math-Sub, but worse when post-trained on MATH, likely due to the larger distribution shift between training dataset and the evaluation one.

\begin{figure}[h!]
    \centering
    \begin{subfigure}[t]{0.4\linewidth}
        \centering
        \includegraphics[width=\linewidth]{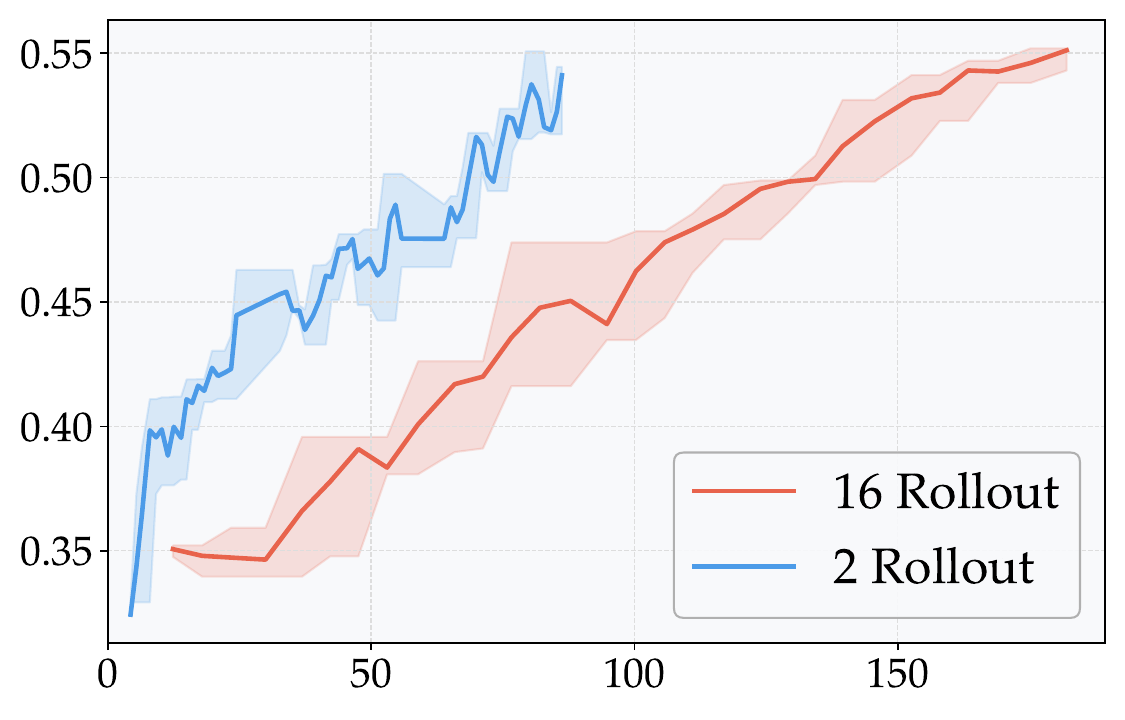}
        \caption{Train Reward ($y$) vs. Time ($x$) [Geo3K]}
        \label{fig:geo3k-train}
    \end{subfigure}
    \begin{subfigure}[t]{0.4\linewidth}
        \centering
        \includegraphics[width=\linewidth]{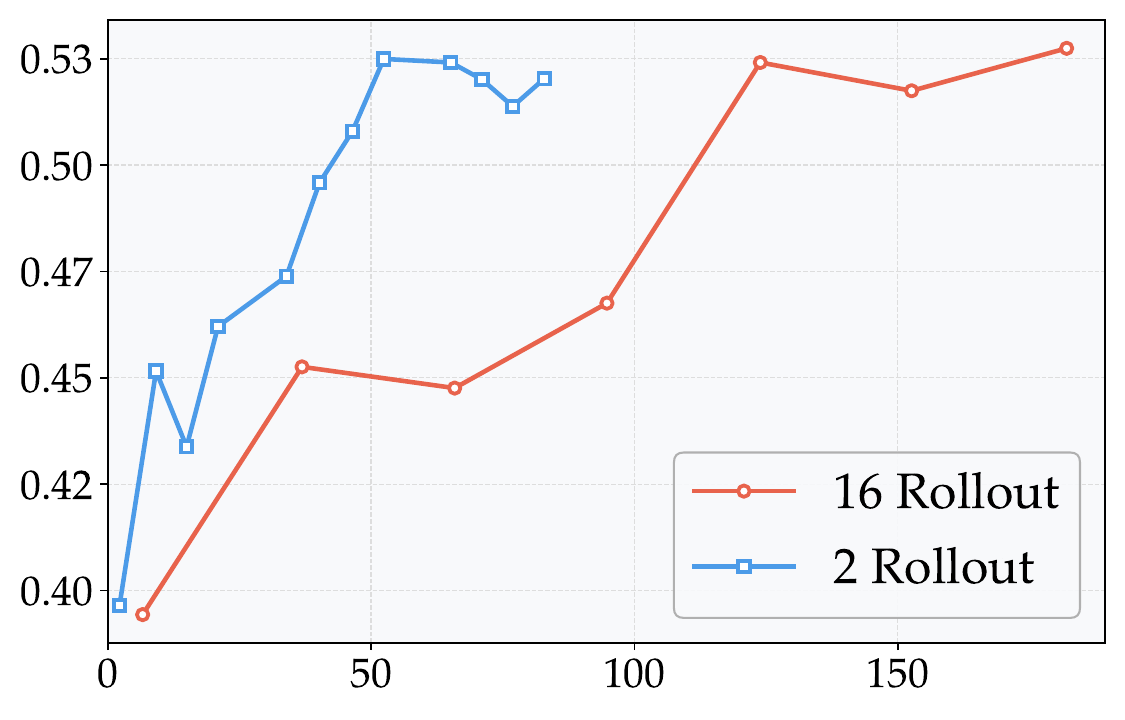}
        \caption{Test Acc. ($y$) vs. Time ($x$) [Geo3K]}
        \label{fig:geo3k-test}
    \end{subfigure}\hfill
    \begin{subfigure}[t]{0.4\linewidth}
        \centering
        \includegraphics[width=\linewidth]{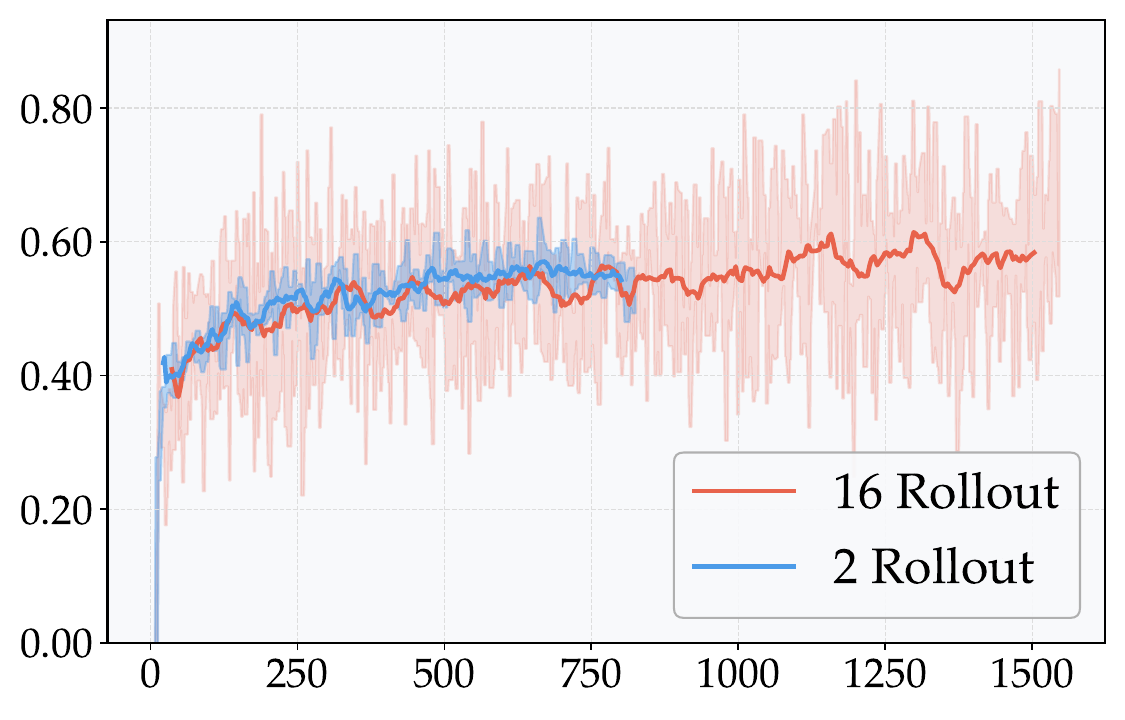}
        \caption{Train Reward ($y$) vs. Time ($x$) [CodeR1]}
        \label{fig:code-r1-train}
    \end{subfigure}
    \begin{subfigure}[t]{0.4\linewidth}
        \centering
        \includegraphics[width=\linewidth]{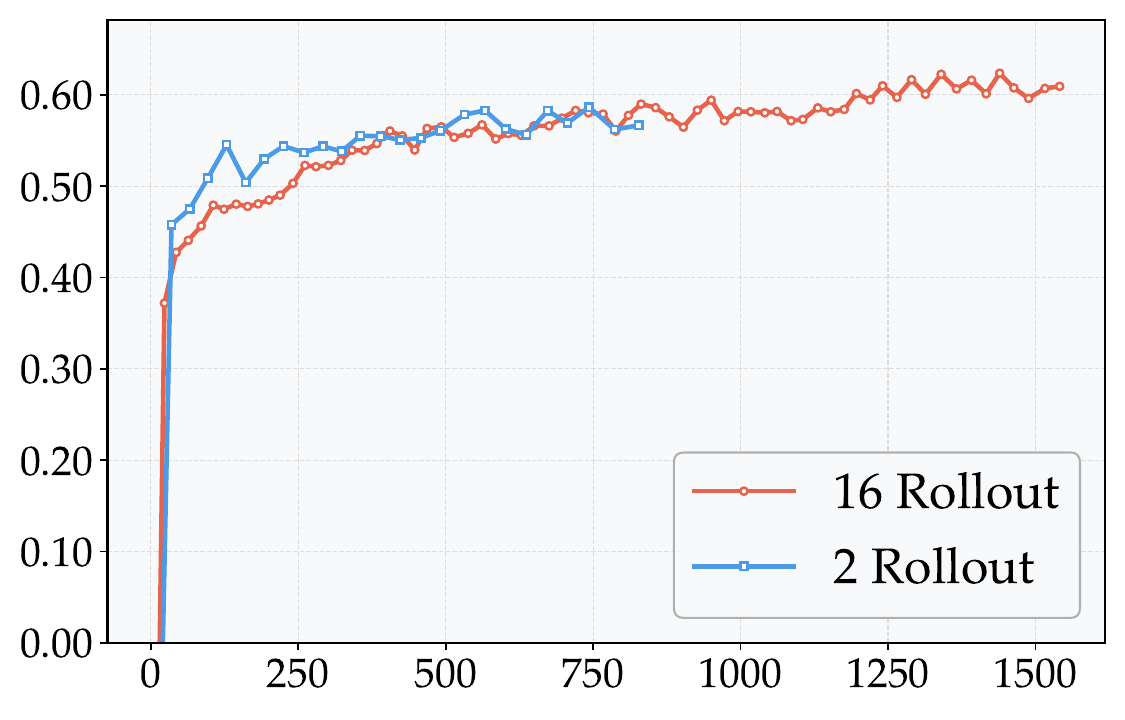}
        \caption{Test Acc. ($y$) vs. Time ($x$) [CodeR1]}
        \label{fig:code-r1-test}
    \end{subfigure}

    \caption{The performance of 2-GRPO over training time (mins) on Geometry3K and Code-R1.}
    \label{fig:vision_and_code_benchmark}
\end{figure}

We extend the evaluation of 2-GRPO to additional RLVR tasks beyond mathematical reasoning, including Vision Reasoning (Geometry3K) and Code Generation (Code-R1), with results reported in Figure~\ref{fig:vision_and_code_benchmark}. The results demonstrate that 2-GRPO remains both effective and efficient across these diverse tasks, highlighting its broader applicability beyond math reasoning.
As shown in the figure, 2-GRPO converges substantially faster than 16-GRPO, owing to the reduced number of samples generated and updated per step. This phenomenon is consistent with our theoretical findings, which identify the role of the group as providing contrastive sample pairs. Reducing the group size not only preserves performance but also accelerates the learning process.

\subsection{Ablation Study: The Effect of Group Size}
\begin{figure}
    \centering
    \includegraphics[width=\textwidth]{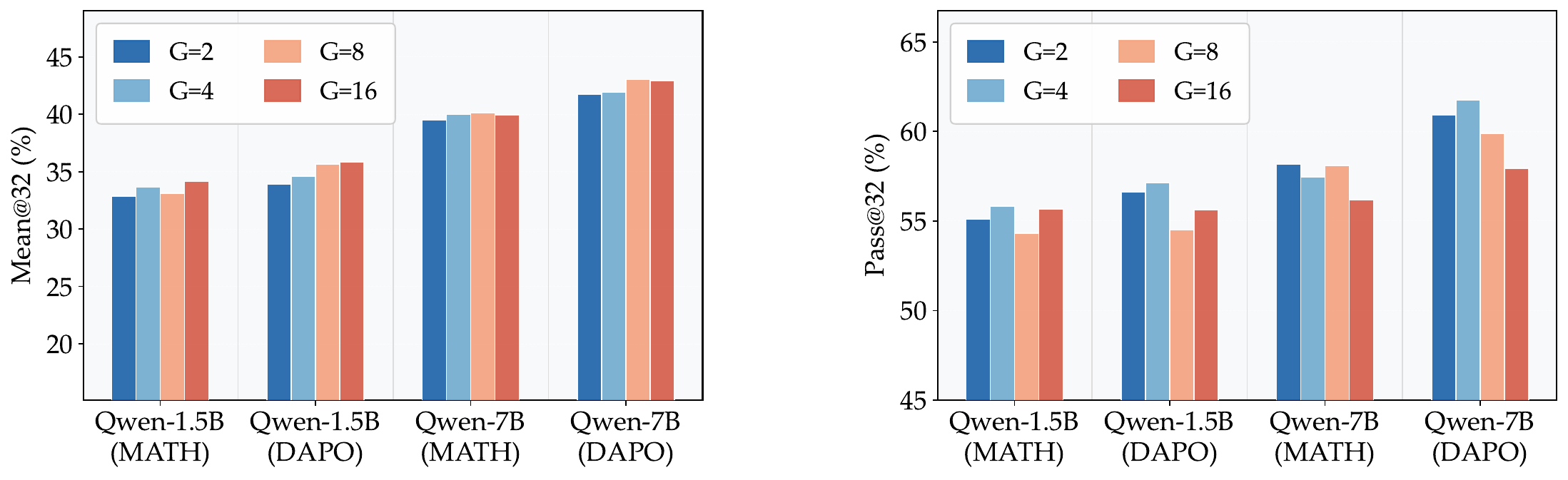}
    \caption{Average performance evaluated over $5$ Math benchmarks with group size $G=2,4,8,16$. The results are reported in  Mean@$32$ (left) and Pass@$32$ (right).}
    \label{fig:group_size}
\end{figure}
Our proposed 2-GRPO changes the group size while also adjusting the training batch size 
and the learning rate to account for the reduced number of rollouts per prompt (discussed in Appx.~\ref{appx:var_reduction}).
To isolate the effect of group size, we conduct an ablation study over different group sizes using the exact same configuration: 10 training epochs, a generation batch size of 512 prompts, a training batch size of 32 prompts, and a learning rate of $10^{-6}$.

It is worth noting that this setting is slightly unfavorable to smaller group sizes -- the actual training mini-batch size by \# rollouts is \# prompts per batch multiplied by the group size. 
Therefore, GRPO with smaller group sizes in the ablation study suffered from higher variance of gradient estimates (see~\ref{appx:gradient_estimate} for details).
Nonetheless, Figure~\ref{fig:group_size} shows that the Mean@$32$ differences among $G=2,4,8,16$ are consistently small across all settings.
Moreover, increasing the group size does not reliably improve Pass@$32$: larger groups do not consistently outperform smaller ones, and in some cases smaller groups achieve better Pass@$32$.
The full results of the ablation study are provided in Table~\ref{tab:ablation} in Appx.~\ref{appx:ablation_group_size}.




\section{Conclusion}

In this work, we demonstrate that GRPO \textit{de facto} functions as contrastive learning. 
We argue that the primary role of the group mechanism is not for accurate value-baseline estimation, as commonly assumed, but for the efficient construction of contrastive signals.
Based on this insight, we reveal the fundamental connection between GRPO and DPO—they are two echoes of the same contrastive gradient optimization principle, reflected through the online and offline RL settings, respectively.
To further validate this insight, we introduce 2-GRPO, a minimal variant with only two rollouts per prompt. 
Although this setting is degenerate from the standpoint of traditional advantage estimation, 
it remains theoretically well motivated under our contrastive framework.
Empirically, 2-GRPO achieves performance comparable to 16-GRPO while substantially reducing the computational overhead of rollout generation and policy optimization. 
These results support our hypothesis and suggest a more efficient design principle for RL algorithms for LLMs. More broadly, while our analysis focuses on GRPO, the insights developed here may extend to a wider class of group-based RL algorithms.

\newpage
\bibliography{ref}
\bibliographystyle{plain}

\newpage
\appendix
\section*{Appendix}
\section{Related Work}
\label{appx:related_work}
\subsection{Contrastive Learning and LLM Alignment}
Contrastive learning is the cornerstone of self-supervised representation learning \cite{wang2020understanding, he2020momentum, chen2020simple, hu-etal-2022-momentum, wu2024unifying}.
The fundamental objective is to minimize the distance between anchor and positive samples in the representation space while maximizing the distance between the anchor and negative samples.
Given this contrastive nature, the framework shares structural similarity with DPO, which conducts preference learning by increasing the likelihood of preferred completions relative to dispreferred ones.
While recent literature explores the theoretical connections between RLHF and contrastive learning \cite{hejna2023contrastive, flet2024contrastive, lv2025hidden}, our work establishes a formal link between GRPO and DPO through a contrastive lens.
This provides a unified analytical framework for understanding alignment.
Specifically, we attribute the efficacy of GRPO to the construction of contrastive pairs, which serves as a control variate to reduce the variance of the gradient estimator.
This analysis offers generalizable insights to broader alignment algorithms.

\subsection{Adaptive Rollouts in RLVR}
RL post-training has demonstrated significant success in enhancing  LLM performance across diverse domains \cite{wu2025advancing, zhang2025rearank}.
Unlike SFT, RL requires the model to generate online samples during training.
Although modern frameworks integrate high-throughput inference engines such as vLLM and SGLang, the autoregressive nature of LLMs ensures that the generation phase remains a primary computational bottleneck.
This challenge is exacerbated by the common intuition that LLM-based RL often necessitates large group sizes to achieve good performance.
To mitigate this overhead, recent studies have proposed selective or adaptive sampling techniques to reduce the number of rollouts without compromising performance \cite{zheng2025act, zhang2025speed, zhu2025shuffle}.
Within this context, 2-GRPO serves as a robust baseline.
Furthermore, our contrastive analysis of GRPO opens a new design space for developing efficient sampling algorithms in RLVR.
\section{Theorems}

\subsection{Variance Reduction of Policy Gradient Estimate}
\label{appx:var_reduction}

Given a prompt, consider the random variable (r.v.) of the reward $r$ (which can be replaced by the advantage $a$) and the r.v. of the policy gradient $\pg$ (corresponding to $\frac{1}{|o_i|}\sum_{t=0}^{|o_i|} \nabla_\theta \log\llm (o_{i,t} | o_{i,<t}, q)$ in VPG or $\frac{1}{|o_i|}\sum_t^{|o_i|} \nabla_\theta \rho_{i,t}$ in PPO/GRPO).
Since $\E[\pg]=0$ over all potential actions, the variance of the product of these r.v.'s can be written as:
\begin{equation}
\begin{aligned}
    \text{Var}(r \cdot \pg) &=\text{Var}(\pg) \left[ \text{Var}(r) + (\E[r])^2 \right] + \underbrace{\text{Cov}(r^2, \pg^2) - (\text{Cov}(r, \pg))^2}_{\text{Interaction term}}\,.
\end{aligned}
\label{eq:var_reward_grad}
\end{equation}
The interaction term can be ignored when importance sampling and clipping are applied,
as the gradient is bounded in a small region.
Previous work~\cite{baird1993advantage, schulman2015trust, schulman2015high, schulman2017proximal} shows that replacing raw rewards with advantage functions ($\E[a]=0$) effectively reduces variance, leading to more stable and improved RL optimization.

\subsection{Mean Estimation with Samples ${n=2}$}
\label{appx:student}

The instability of normalization with extremely small samples is a well-documented phenomenon in classical statistics, dating back to the seminal work of \textit{William Sealy Gosset} (published under the pen name \textit{Student}) ~\cite{student1908probable}. 
For a sample size of $n=2$, 
the degrees of freedom $df=1$ result in a normalization factor that follows a Cauchy distribution. 
Such small-sample estimates of variance are highly skewed, leading to normalized outputs with infinite variance and no defined mean, undermining the goal of statistical stability.

\subsection{Reveal GRPO as Contrastive Learning}
\label{appx:rewrite_grpo}

\begin{proof}[Proof of Proposition~\ref{prop:grpo_is_cl}]

In the RLVR setting, rewards are binary, which leads to binary advantages given a prompt.
Let $A^+_q, A^-_q$ denote the positive and negative advantage, respectively.
From \eqref{eq:grpo_adv}, we can have 
\begin{equation}
    \begin{aligned}
A^+_q &= \frac{1-\hat{p}_q}{\sqrt{\hat{p}_q(1-\hat{p}_q)}}= \sqrt{\frac{1-\hat{p}_q}{\hat{p}_q}} \;, \\
A^-_q &= \frac{0-\hat{p}_q}{\sqrt{\hat{p}_q(1-\hat{p}_q)}}= -\sqrt{\frac{\hat{p}_q}{1-\hat{p}_q}} \;.
    \end{aligned}
\end{equation}

The clipping function is
\begin{equation}
    \text{clip}(x, 1-\epsilon,1+\epsilon) =
    \begin{cases}
    x, & |x-1| \leq \epsilon \\
    1-\epsilon, & x < 1-\epsilon \\
    1+\epsilon, & x > 1+\epsilon
    \end{cases}\, ,
\end{equation}
which means that $x$ will be assigned to $1-\epsilon$ ($1+\epsilon$) if $x$ is less (greater) than $1-\epsilon$ ($1+\epsilon$).
For simplifying notation, let $\clip^+(x)=\min[x, 1+\epsilon]$ and $\clip^-=\max[x, 1-\epsilon]$.

The key derivation of rewriting GRPO objective is as follows:
\begin{equation}
   \begin{aligned} 
    &\mathcal{J}_{\text{GRPO}}(\theta) \\
    &= \mathbb{E}_{\substack{q \sim \mathcal{Q} \\ \{o_i\}_{i=1}^G \sim \pi_{\theta_\text{old}}(\cdot|q)}}  \frac{1}{G}\sum_{i=1}^G \frac{1}{|o_i|} \sum_{t=1}^{|o_i|} \clip \left( \frac{\llm (o_{i,t} |o_{i,<t}, q )}{\pi_{\theta_{\text{old}}} (o_{i,t} |o_{i,<t}, q )} A_{i,t} \right)  \, , \\
    &= \mathbb{E}_{\substack{q \sim \mathcal{Q} \\ \{o_j\}_{j=1}^{G^+} \sim \pi_{\theta_{\text{old}}}^+(\cdot|q) \\ \{o_k\}_{k=1}^{G^-} \sim \pi_{\theta_{\text{old}}}^-(\cdot|q)}} \\
    &\frac{1}{G} \left( \sum_{j=1}^{G^+} \frac{1}{|o_j|} \sum_{t=1}^{|o_j|} A_{j}^+ \clip^+ \left( \frac{\llm (o_{j,t} |o_{j,<t}, q )}{\pi_{\theta_{\text{old}}} (o_{j,t} |o_{j,<t}, q )} \right) +  \sum_{k=1}^{G^-} \frac{1}{|o_k|} \sum_{t=1}^{|o_k|} A_{k}^- \clip^- \left( \frac{\llm (o_{k,t} |o_{k,<t}, q )}{\pi_{\theta_{\text{old}}} (o_{k,t} |o_{k,<t}, q )} \right) \right)  \, ,\\
    &= \mathbb{E}_{\substack{q \sim \mathcal{Q} \\ \{o_j\}_{j=1}^{G^+} \sim \pi_{\theta_{\text{old}}}^+(\cdot|q) \\ \{o_k\}_{k=1}^{G^-} \sim \pi_{\theta_{\text{old}}}^-(\cdot|q)}}\\
    &A^+_q \frac{G^+}{G} \frac{1}{G^+}\sum_{j=1}^{G^+}  \frac{1}{|o_j|} \sum_{t=1}^{|o_j|} \clip^+ \left( \frac{\llm (o_{j,t} |o_{j,<t}, q )}{\pi_{\theta_{\text{old}}} (o_{j,t} |o_{j,<t}, q )} \right)+ A^-_q\frac{G^-}{G} \frac{1}{G^-}  \sum_{k=1}^{G^-} \frac{1}{|o_k|} \sum_{t=1}^{|o_k|} \clip^- \left( \frac{\llm (o_{k,t} |o_{k,<t}, q )}{\pi_{\theta_{\text{old}}} (o_{k,t} |o_{k,<t}, q )} \right) \, ,\\
    &=\mathbb{E}_{\substack{q \sim \mathcal{Q} \\  \{o_j\}_{j=1}^{G^+} \sim \pi_{\theta_{\text{old}}}^+(\cdot|q) \\ \{o_k\}_{k=1}^{G^-} \sim \pi_{\theta_{\text{old}}}^-(\cdot|q)}}\\
    &\sqrt{\widehat{\Var}_G(q) } \left( \frac{1}{G^+}\sum_{j=1}^{G^+}  \frac{1}{|o_j|}\sum_{t=1}^{|o_j|} \clip^+ \left(  \frac{\llm (o_{j,t} |o_{j,<t}, q )}{\pi_{\theta_{\text{old}}} (o_{j,t} |o_{j,<t}, q )} \right) - \frac{1}{G^-} \sum_{k=1}^{G^-} \frac{1}{|o_k|} \sum_{t=1}^{|o_k|} \clip^- \left( \frac{\llm (o_{k,t} |o_{k,<t}, q )}{\pi_{\theta_{\text{old}}} (o_{k,t} |o_{k,<t}, q )} \right) 
    \right) \, .
   \end{aligned} 
\end{equation}
The second equation is obtained by dividing the trajectories into two groups: positive and negative.
The third equation is obtained by the fact that all positive advantages are the same and that all negative advantages are the same.
Since $A^+ \frac{G^+}{G}=\sqrt{\frac{1-\hat{p}}{\hat{p}}} \hat{p}=\sqrt{(1-\hat{p})\hat{p}}$ and $A^- \frac{G^-}{G}= -\sqrt{(1-\hat{p})\hat{p}}$, we obtain~\eqref{eq:grpo_finite}.
When $G \to \infty$, we have the following facts:
\begin{equation}
    \begin{aligned}
    &\lim_{G \to \infty} G^+ = \infty \;,\\
    &\lim_{G \to \infty} G^- = \infty \\
    &\lim_{G \to \infty} \sqrt{(1-\hat{p})\hat{p}} = \sqrt{(1-p)p} \;,\\
    &\lim_{G^+ \to \infty} \frac{1}{G^+}\sum_{j=1}^{G^+} f(o_j) = \E_{o_j \sim O_\theta^+} f(o_j) \;,\\
    &\lim_{G^- \to \infty} \frac{1}{G^-}\sum_{k=1}^{G^-} f(o_k) = \E_{o_k \sim O_\theta^-} f(o_k) \;.\\
    \end{aligned}
\end{equation}

Then the GRPO objective has the following gradient w.r.t. parameter $\theta$:
\begin{align}
&\nabla_\theta \mathcal{J}_{\text{GRPO}} = \nonumber \\& \underset{q \sim \mathcal{Q}}{\E}\sqrt{\widehat{\Var}_G(q)}  
\Bigg(
\frac{1}{G_q^+}\sum_{j=1}^{G_q^+} \sum_t^{|o_j^+|} \frac{\mathbbm{1}^\epsilon_{j,t} \nabla_\theta\pi_\theta (o_{j,t}^+|o_{j,<t}^+,q)}{|o_j^+|\pi_{\theta_{\text{old}}}(o_{j,t}^+|o_{j,<t}^+,q)}
- \frac{1}{G_q^-} \sum_{k=1}^{G_q^-} \sum_t^{o_k^-} \frac{\mathbbm{1}^\epsilon_{k,t}\nabla_\theta\pi_\theta (o_{k,t}^-|o_{k,<t}^-,q)}{|o_k^-|\pi_{\theta_{\text{old}}}(o_{k,t}^-|o_{k,<t}^-,q)}
\Bigg)  
\\ & =
\label{eq:grpo_gradient}
\underset{q \sim \mathcal{Q}}{\E} \Bigg[
\underbrace{\frac{1}{G_q^+}\sum_{j=1}^{G_q^+} \sum_t^{|o_j^+|} c(o_{j,t}^+|o_{j,<t}^+,q) \nabla_\theta\pi_\theta (o_{j,t}^+|o_{j,<t}^+,q)}_{\text{Positive}} 
- \underbrace{\frac{1}{G_q^-} \sum_{k=1}^{G_q^-} \sum_t^{|o_k^-|} c(o_{k,t}^-|o_{k,<t}^-,q) \nabla_\theta\pi_\theta (o_{k,t}^-|o_{k,<t}^-,q)}_{\text{Negative}}
\Bigg]
\end{align}
where $\mathbbm{1}^\epsilon_{j,t}$ is an indicator function if the token $o_{j,t}$ is clipped and $c(o_{i,t}|o_{i,<t},q):=\frac{\sqrt{\widehat{\Var}(q)}\mathbbm{1}^\epsilon_{i,t} }{|o_i|\pi_{\theta_{\text{old}}}(o_{i,t}|o_{i,<t},q)}$.
Compare~\eqref{eq:grpo_gradient} with Def.~\ref{def:cl}, the derivative of GRPO is a Monte Carlo estimator of contrastive derivative.
\end{proof}

\subsection{Further Discussion on Importance Sampling and the Log-likelihood Term}
\label{appx:log_vs_is}

Most autoregressive LLMs adopt causal probability modelling as $\log\llm (o | q) = \sum\log\llm (o_t|o_{<t},q)$. This decomposition leads to the following trajectory-level form to describe the gradient of token probabilities:
\begin{equation}
\nabla_\theta \log\llm(o|q) =  \sum_t \frac{1}{ \llm(o_t|o_{<t},q)}\nabla_\theta \llm (o_t|o_{<t},q) \,.
\end{equation}
DPO follows a similar structural derivation.

It is worth mentioning that the importance sampling in PPO can be viewed as a natural extension of such gradient form for online on/off-policy RL~\cite{schulman2017proximal}. However, the token-level importance sampling in PPO and vanilla GRPO often obscures this direct connection at the trajectory level.

Recent subsequent variants of GRPO~\cite{zheng2025group, zhao2025geometric, pang2025theory}, e.g., GSPO and TIC-GRPO, utilize sequence-level importance sampling. This formulation allows us to draw a direct connection between importance sampling and the log-likelihood terms:
\begin{equation}
\nabla_\theta \frac{\pi_\theta(o \mid q)}{\pi_{\theta_{\mathrm{old}}}(o \mid q)} = \frac{\pi_\theta(o \mid q)}{\pi_{\theta_{\mathrm{old}}}(o \mid q)} \sum_t \frac{1}{\pi_\theta(o_t \mid o_{<t}, q)} \nabla_\theta \pi_\theta(o_t \mid o_{<t}, q) \,.
\end{equation}
It is straightforward to see from the gradient form that the importance sampling term adjusts the Log-likelihood term by a coefficient $\frac{\pi_\theta(o \mid q)}{\pi_{\theta_{\mathrm{old}}}(o|q)}$. The token-level importance sampling in PPO and GRPO behaves similarly by applying token-level correction.

The clipping applied on top of importance sampling is a minor additional modification, which we do not elaborate on here.


\subsection{Proof of Lemma~\ref{lemma:dpo_is_cl}: DPO is 1-vs-1 contrastive learning}
\label{appx:dpo_is_cl}

\begin{lemma}
\label{lemma:dpo_is_cl}
The DPO loss is a $1$-vs-$1$ contrastive loss estimator.
\end{lemma}

\begin{proof}[Proof of Lemma~\ref{lemma:dpo_is_cl}]
\label{proof: dpo_is_cl}
The DPO loss (\eqref{eq:dpo}) has the following derivatives:
\begin{equation}
\begin{aligned}
\label{eq:dpo_derivative}
&\nabla_\theta \mathcal{L}_{\text{DPO}} =  -\beta \underset{[(q, o^+, o^-) \sim \gD_\text{DPO}]}{\E}  \Big [
\sigma(\hat{r}_\theta (q,o^-) - \hat{r}_\theta (q,o^+))  \left(
\nabla_\theta \log \llm (o^+|q)  - \nabla_\theta \log \llm (o^-|q)
\right)
\Big ]  \; \,\\
&=- \underset{[(q, o^+, o^-) \sim \gD_\text{DPO}]}{\E} \Bigg[
\underbrace{\sum_{t}^{|o^+|} c(o^+_t|o^+_{<t},q) \nabla_\theta \pi_\theta(o^+_t |o^+_{<t},q )}_{\text{Positive}} - \underbrace{\sum_{t}^{|o^-|} c(o^-_t|o^-_{<t},q) \nabla_\theta \pi_\theta(o^-_t |o^-_{<t},q )}_{\text{Negative}}
\Bigg]
\end{aligned}
\end{equation}
where $\hat{r}_\theta = \beta(x,y) \log \frac{\llm (y|x)}{\pi_{\text{ref}}(y|x)}$; $\sigma$ denotes the sigmoid function;
and $c(o_t|o_{<t},q):=\frac{\beta\sigma(\hat{r}_\theta (q,o^-) - \hat{r}_\theta (q,o^+))}{\pi_\theta(o|q)}$ aligning with Def.~\ref{def:cl}.
\end{proof}

\subsection{GRPO v.s. DPO from Contrastive Learning}
\label{appx:grpo_vs_dpo}

As shown in Appx.~\ref{appx:rewrite_grpo} and ~\ref{appx:dpo_is_cl}, GRPO and DPO admit the following gradient formulations:
\begin{align}
&\nabla_\theta \mathcal{J}_{\text{GRPO}} = \underset{q \sim \mathcal{Q}}{\mathbb{E}} \Bigg[ \underbrace{\sum_t^{|o^+|} c(o_{t}^+|o_{<t}^+,q) \nabla_\theta\pi_\theta (o_{t}^+|o_{<t}^+,q)}_{\text{Positive}} - \underbrace{\sum_t^{|o_k^-|} c(o_{k,t}^-|o_{k,<t}^-,q) \nabla_\theta\pi_\theta (o_{k,t}^-|o_{k,<t}^-,q)}_{\text{Negative}} \Bigg] \\
&\nabla_\theta \mathcal{L}_{\text{DPO}} = - \underset{[(q, o^+, o^-) \sim \mathcal{D}_\text{DPO}]}{\mathbb{E}} \Bigg[ \underbrace{\sum_{t}^{|o^+|} c(o^+_t|o^+_{<t},q) \nabla_\theta \pi_\theta(o^+_t |o^+_{<t},q )}_{\text{Positive}} - \underbrace{\sum_{t}^{|o^-|} c(o^-_t|o^-_{<t},q) \nabla_\theta \pi_\theta(o^-_t |o^-_{<t},q )}_{\text{Negative}} \Bigg]
\end{align}

These expressions reveal that both maximizing the GRPO objective and minimizing the DPO loss correspond to the same underlying contrastive learning mechanism, differing only in the specific design of the coefficient $c(o_t^+|o_{<t}^+,q)$.

The key distinction lies in how the coefficient $c(\cdot)$ is instantiated under different RL settings (online vs.\ offline):
\begin{itemize}[leftmargin=*, itemsep=0pt, topsep=0pt, parsep=0pt, partopsep=0pt]
\item \textbf{Importance sampling term} (online GRPO) vs.\ \textbf{log-likelihood term} (offline DPO) (see Appx.~\ref{appx:log_vs_is} for detailed discussion).
\item \textbf{Reference-model regularization}: an explicit KL term (GRPO) vs.\ implicit incorporation into the ``advantage'' (DPO).
\end{itemize}

Importantly, these differences do not alter the fundamental optimization structure, but rather reflect distinct design choices tailored to their respective RL regimes.

\subsection{Proof of Proposition~\ref{prop:control_variate}}
\label{proof:control_variate}
\begin{proof}

\begin{equation}
\begin{aligned}
\Var(\bm{g}^+ - c\bm{g}^-) &= \Var(\bm{g}^+) + c^2\Var(\bm{g}^-) -2c\Cov(\bm{g}^+, \bm{g}^-) \, ,  \\
&= \Var(\bm{g}^+) - \frac{\Cov^2(\bm{g}^+, \bm{g}^-)}{\Var(\bm{g}^-)} \, , \\
&= (1-\rho^2)\Var(\bm{g}^+) \, .
\end{aligned}
\end{equation}
The first equation is obtained by the definition of variance.
The second equation is obtained by substituting $c=\frac{\Cov(\bm{g}^+, \bm{g}^-)}{\Var(\bm{g}^-)}$.
The third equation is hold because $\rho = \frac{\Cov(\bm{g}^+ , \bm{g}^-)}{\sqrt{\Var(\bm{g}^+) \Var(\bm{g}^-)}}$.
On the other hand, consider $f(c) = c^2\Var(\bm{g}^-) -2c\Cov(\bm{g}^+, \bm{g}^-)$.
If $0 \leq c \leq 2\frac{\Cov^2(\bm{g}^+, \bm{g}^-)}{\Var(\bm{g}^-)} $, then $f(c) \leq 0$.
\end{proof}

\subsection{The Correlation between The Positive and The Negative}
\label{appx:correlation_pos_neg}
We do not have access to the joint distribution of positive and negative gradients, so direct empirical estimation of their covariance is infeasible.
Instead, we use the law of total covariance:
\begin{equation}
\mathrm{Cov}(\bm{g}^+ , \bm{g}^-) = \mathbb{E}_p[\mathrm{Cov}(\bm{g}^+, \bm{g}^- \mid p)] + \mathrm{Cov}_p(\mathbb{E}[\bm{g}^+|p], \mathbb{E}[\bm{g}^-|p])\, .
\end{equation}
Under conditionally independent sampling, the first term is zero, so we estimate the second term across prompts.
Statistics are computed over 100 prompts with 10 responses each on MATH with Qwen-1.5B. As high-dimensional vectors usually have small dot-products, we report a baseline as reference where pos/neg pairs are randomly permuted.
\begin{table}[htbp]
    \centering
    \caption{Covariance metrics between positive and negative gradients.}
    \label{tab:gradient_cov}
    
    \begin{tabular}{lcc}
        \toprule
        \textbf{Metric} & $Cov(g^+, g^-)$ & $Cov_{\mathrm{perm}}$ \\
        \midrule
        Value & 0.08705 & 0.00128 \\
        \bottomrule
    \end{tabular}
\end{table}

As shown in the table, the covariance between positive and negative gradients from the same prompt is significantly larger than that between randomly paired positive and negative gradients, which confirms our assumption.

\subsection{Proof of Proposition~\ref{prop:adv_estimate}}
\begin{proof}
    \label{proof: advantage_estimate}
    \textbf{Case 1.} Notice that $\hat\sigma=\sqrt{\frac{1}{2N}\sum_{k=1}^{2N} (X_k-\hat\mu)^2}
=\sqrt{\hat\mu(1-\hat\mu)}$ and $\hat\mu = \frac{1}{2N}\sum_{k=1}^{2N}X_k$.  Fix an index $i$ and condition on the event $\{X_i=x\}$ with $x\in\{0,1\}$. In this case, by the strong law of large numbers and the continuous mapping theorem, we have $\hat\mu \stackrel{a.s.}{\rightarrow} p$  and $\hat\sigma \stackrel{a.s.}{\rightarrow}\sqrt{p(1-p)}$.
Thus, it follows that
\[
\lim_{\epsilon\rightarrow 0}\,\lim_{N\to\infty}\,\E\!\left[Y_i\mid X_i=x\right]
=\frac{x-p}{\sqrt{p(1-p)}}.
\]

\textbf{Case 2.}
When $X_{i,1} =  X_{i,2}$, we have $X_{i,j}=\hat\mu_i$ and $Y_{i,j}=0$  for any $j \in  \{1,2\}$.  
When $X_{i,1}\neq X_{i,2}$, we have $\hat\mu_i=0.5$, $\hat\sigma_i=0.5$, and $Y_{i,j}=\frac{2X_{i,j}-1}{1+2\epsilon}.$
By the law of total expectation, it follows that
\[
\E\left[Y_{i,j}\mid X_{i,j}=1\right]= \frac{1-p}{1+2\epsilon},\qquad
\E\left[Y_{i,j}\mid X_{i,j}=0\right]=\frac{-p}{1+2\epsilon}.
\]
Thus, we have
\[
\lim_{\epsilon\rightarrow 0}\E\!\left[Y_{i,j}\mid X_{i,j}=x\right]=x-p.
\]
\end{proof}


\subsection{Proof of Lemma~\ref{lemma:var}}
\label{appx:gradient_var_lemma}

\begin{proof}[Proof of Lemma~\ref{lemma:var}]
\label{proof: gradient_var_lemma}
\begin{equation}
\label{eq:var}
\begin{aligned}
\Var(\hat{\pg}_{B})
&= \Var_{\{\vx_i\}_{i=1}^B}
\left( \frac{1}{B} \sum_{i=1}^B \pg(\vx_i) \right) \\
&= \frac{1}{B^2} \sum_{i=1}^B \Var_{\vx_i}\!\left(\pg(\vx_i)\right) = \frac{\Var_{\vx}\!\left(\pg(\vx)\right)}{B} \; .
\end{aligned}
\end{equation}

where the second and third equalities are obtained by the properties of independence and identity in i.i.d. data, respectively. 
By the above equation, increasing $B$ decreases $\Var$.
\end{proof}

\section{Theoretical Analysis of 2-GRPO}

\subsection{Implicit Weighting in Stochastic Optimization}
\label{appx:adv_estimate}

At first glance, 2-GRPO appears to use only fixed advantages, $A^+=1$ and $A^-=-1$, ignoring prompt-level success rates.
However, under mini-batch stochastic optimization, 2-GRPO implicitly reweights prompts through their likelihood of forming contrastive pairs.

Standard GRPO relies on the empirical success rate $\hat{p}_q$ to estimate the true correctness probability $p_q$ for advantage assignment, relying on larger group sizes for accuracy. 
While this mechanism appears degenerate in 2-GRPO, we show that, through the lens of stochastic optimization, 2-GRPO implicitly estimates the advantage.

Moreover, 2-GRPO does not simply estimate the large-group GRPO gradient with fewer samples; it induces a different prompt-level weighting that prioritizes prompts likely to yield contrastive pairs.

\begin{proposition}
\label{prop:adv_estimate}
    Given a constant $p \in (0,1)$ and a small positive constant $\epsilon$, we consider two scenarios below:
\begin{itemize}[left=0pt, nosep]
    \item \textbf{Case 1}:
    Consider $X_1, \cdots, X_{2N} \stackrel{\text{i.i.d.}}{\sim}  \text{Bernoulli}(p)$. Let $Y_i = \frac{X_i - \hat{\mu}}{\hat{\sigma} + \epsilon}$, where $\hat\mu = \frac{1}{2N}\sum_{i=1}^{2N}X_i$ and $\hat\sigma = \sqrt{\frac{1}{2N}\sum_{i=1}^{2N}\left(X_i-\hat{\mu}\right)^2}$. Then, it follows that
     \begin{equation}
     \lim_{\epsilon \rightarrow 0}\lim_{N\rightarrow \infty} \E[Y_i| X_i=x]=\frac{x-p}{\sqrt{p(1-p)}}. 
     \end{equation}
     
    \item \textbf{Case 2}:
        Consider $N$ pairs of $(X_{i,1}, X_{i,2})$ with each $X_{i,j} \stackrel{\text{i.i.d.}}{\sim}  \text{Bernoulli}(p)$. Let $Y_{i,j} = \frac{X_{i,j} - \hat{\mu}_i}{\hat{\sigma}_i + \epsilon}$, where 
    $\hat\mu_i=\frac{1}{2}(X_{i,1}+X_{i,2})$ and $\hat\sigma_i=\sqrt{\frac{1}{2}\sum_{j=1}^2(X_{i,j}-\hat\mu_i)^2}$. Then, it follows that 
        \begin{equation}
    \lim_{\epsilon \rightarrow 0}\lim_{N\to\infty} \E[Y_{i,j}| X_{i,j}=x]= x-p.
         \end{equation}
\end{itemize}
Term $\lim_{\epsilon \rightarrow 0, N\to \infty}\E[Y_{i,j}| X_{i,j}=x]$ differs from $\lim_{\epsilon \rightarrow 0,N\rightarrow \infty}\E[Y_i| X_i = x]$ by a scaling factor $\frac{1}{\sqrt{p(1-p)}}$.

\end{proposition}

In Proposition~\ref{prop:adv_estimate} (proof in Appx.~\ref{proof: advantage_estimate}), \textbf{Case 1} corresponds to regular GRPO with sufficiently large group size. In this case, 
$\E[Y_i| X_i=1]$ and $\E[Y_i| X_i=0]$ are, respectively, the advantage estimates of positive and negative trajectories given a prompt, dependent on the success probability $p_q$.
A large $G$ will lead to a better estimate of the success probability $p_q$.
\textbf{Case 2} corresponds to 2-GRPO, where $\E[Y_{i,j} |X_{i,j} = 1]$ and $\E[Y_{i,j} |X_{i,j} = 0]$ are advantage estimates, which are also dependent on the success rate $p_q$, amortizing over multiple stochastic updates.

2-GRPO produces advantage estimates that differ from standard GRPO solely by a scaling factor; 
this factor is effectively a design choice. 
Whether such a scaling is beneficial remains an open question~\cite{li2025DiscoReinforcingLarge}.

\subsection{Key of Variance Reduction: the Training Batch Size, not the Group Size}
\label{appx:gradient_estimate}

Beyond the inherent variance reduction mechanisms of PPO and GRPO, it is generally understood that using a larger group of rollouts yields a lower-variance policy gradient estimate. 
However, this perspective overlooks the practicalities of mini-batch optimization. 
In this section, we analyze the practical gradient variance within a mini-batch setting. 
To facilitate this discussion, we focus strictly on the optimization phase and treat the sampled rollouts as fixed training data for notational simplicity.

Note that there are two notions of ``batch size'' in \textit{VERL}: \texttt{data.train\_batch\_size} denotes the rollout-generation batch size (by \# prompts), whereas \texttt{actor.ppo\_mini\_batch\_size} denotes the optimization mini-batch size (by \# prompts). 
However, the effective number of samples during optimization is actually \texttt{actor.ppo\_mini\_batch\_size * rollout.n}, counted by the number of rollouts.

\begin{definition}[Variance of Gradient Estimate in Mini-Batch]
\label{def:variance}
Without loss of generality, let $\{\vx_i\}_{i=1}^B$ be a batch of $B$ random variables (r.v.'s), where each $\vx_i$ is i.i.d. $\vx\sim\gD$, 
and let $\pg(\vx_i) = \nabla_\theta L_\theta(\vx_i)$ denote the gradient of $L_{\theta}(\vx_i)$ w.r.t. $\theta$.
Define the empirical batch gradient $\hat{\pg}_B = \frac{1}{B} \sum_{i=1}^B \pg(\vx_i)$. 
Note that $\pg(\vx_i)$ and $\hat{\pg}_B$ are dependent r.v.'s of $\boldsymbol{x}_i$ and $\{\boldsymbol{x}_i\}_{i=1}^B$, respectively.
We denote the expectation of the gradient $\bar{\pg} = \mathbb{E}_{\vx\sim\gD}[\pg(\vx)]$. 
The variance of the gradient estimate over the batch is then defined as:
\begin{equation}
{\Var}(\hat{\pg}_B) = 
  {\Var}_{\{\vx_i\}_i^B}(\hat{\pg}_B) = \E_{\{\vx_i\}_i^B} \Big((\hat{\pg}_B - \Bar{\pg})^2\Big). 
\end{equation}
\end{definition}

Following the definition of \emph{Variance of Gradient Estimate in Mini-batch} (Def.~\ref{def:variance}), we provide a lemma for its relationship to the mini-batch size.

\begin{lemma}
\label{lemma:var}
Let $\{ \vx_i\}_{i=1}^{B_1}, \{ \vx_i\}_{i=1}^{B_2}$ be two batches of $B_1$ and $B_2$ r.v.'s, respectively.
Let $\hat{g}_{B_1}, \hat{g}_{B_2}$ denote the empirical batch gradients of these two batches, respectively.
If $B_1 < B_2$, then $\Var[\hat{g}_{B_1}] > \Var[\hat{g}_{B_2}]$.
\end{lemma}

While decreasing the group size in~\eqref{eq:grpo_finite} appears to increase the gradient variance for each individual prompt, 
this conclusion overlooks the total number of rollouts optimized across all prompts in a mini-batch, which is the effective number of examples for optimization.
In Lemma~\ref{lemma:var} (proof in Appx.~\ref{appx:gradient_var_lemma}), we show that a larger batch size $B$ naturally leads to a lower variance of the gradient. 
Note that $B$ is the \textbf{number of rollouts} in each mini-batch rather than the \textbf{number of prompts}.

The actual calculation of GRPO is:
\begin{equation}
\widehat{\mathcal{J}}_{\text{GRPO}}(\theta, G, Q) = \frac{1}{QG}\sum_{j=1}^Q\sum_{i=1}^G A_{ij}\llm^{\text{GRPO}}(o_{ij}|q_j),
\end{equation}
where $$\pi_\theta^{\text{GRPO}}(o|q)=\frac{1}{G} \sum_{i=1}^G \frac{1}{|o_i|}\sum_{t=1}^{|o_i|} \clip \left(A_{i,t}\frac{\llm (o_{i,t}|o_{i,<t},q)}{\pi_{\theta_{\text{old}}}(o_{i,t}|o_{i,<t},q)} \right)$$ 
and $Q$ is the number of prompts in the mini-batch, and the batch size w.r.t the number of rollouts is $B=QG$.
When we decrease $G$, we can increase $Q$ to compensate to retain the same $B$ in a mini-batch.
Since the total number of prompts in the dataset is fixed, 
increasing $Q$ does not increase the total computational cost per training epoch.

\subsection{Exploration on Hard Questions}
\label{appx:breakdown}

A difficult question often requires many attempts to yield a correct answer, which is necessary to form a valid contrastive signal. 
With a smaller group, the likelihood of sampling a correct response in a single iteration may appear lower, potentially raising concerns about degraded learning.


Under a fixed computational budget, 2-GRPO and 16-GRPO explore approximately the same total number of rollouts across all training epochs -- the overall probability of sampling a correct answer under 2-GRPO is not lower than 16-GRPO,
according to the Proposition~\ref{prop:hard_question}.

\begin{proposition}
    \label{prop:hard_question}
        Let $p_i \in [0,1]$ denote the probability that a single rollout under the policy $\pi_i$ produces a correct answer. 
        Then:
        \begin{enumerate}[leftmargin=*]
            \item The probability of obtaining at least one correct answer in $2m$ independent rollouts with policy $\pi_0$ is
            \begin{equation}
                P_{2m} = 1 - (1-p_0)^{2m}.
            \end{equation}
            \item The probability of obtaining at least one correct answer when performing $m$ consecutive  trials of $2$ independent rollouts each, with the corresponding policy $[\pi_0, \pi_1, \cdots, \pi_{m-1}]$ is
            \begin{equation}
                P_{m\times 2} = 1 - \prod_{i=0, \cdots m-1} (1-p_i)^2 \geq 1 - (1-p_0)^{2m} = P_{2m}
                \end{equation}
            when we have $p_{i} \geq p_0, \forall i>0$.
        \end{enumerate}
        Note that the assumption $p_{i} \geq p_0, \forall i>0$ is prevailing,
        as we assume that the reasoning ability of LLM can be improved by RL post-training.
        \end{proposition}

Proposition~\ref{prop:hard_question} suggests that for difficult questions, 2-GRPO does not degrade in effectiveness compared to 16-GRPO given the same budget of the total number of rollouts in whole training process.
Notably, due to its higher frequency of policy updates, 2-GRPO may yield a higher probability of generating correct outputs for hard questions. It is also more adaptive, allowing it to capture nuanced update requirements for varying inputs. This observation also extends to  PPO with the standard single-rollout implementations per epoch against multi-rollout variants.

\section{Experiments}

\subsection{Experiment Details}
\label{appx:exp_details}

\paragraph{Dataset and Baselines}
For math reasoning task, following prior work~\cite{chu2025gpg}, we employ Qwen2.5-Math-1.5B (Qwen-1.5B) and Qwen2.5-Math-7B (Qwen-7B)~\cite{qwen2025Qwen25TechnicalReport} as base models. Both models are post-trained via RL on the MATH~\cite{hendrycksmath2021} and DAPO-Math-17k~\cite{yu2025dapo} datasets, and evaluated on MATH-500~\cite{hendrycksmath2021}, AMC23, Minerva Math~\cite{lewkowycz2022solving}, AIME-2025, and OlympiadBench~\cite{he2024olympiadbench}.
For DAPO-Math-17k dataset, we randomly sample 7.5k questions from the original data to form a subset for training in order to align with the size of MATH.
In addition, we assess the proposed method on DeepSeek-R1-Distill-Qwen-1.5B (DS-1.5B)~\cite{deepseekai2025deepseekr1incentivizingreasoningcapability}, which is post-trained on MATH. Owing to computational constraints, we do not extend its post-training to DAPO-Math-17k.
All 1.5B models are trained on 4 GPUs with 140GB Memory.
Qwen-7B is trained on 8 GPUs with 140GB Memory.
We evaluate model performance using two metrics: Mean@32, the average accuracy across 32 i.i.d. samples, and Pass@32, which measures whether a problem is solved in at least one of those 32 attempts.

For visual reasoning task, we use EasyR1~\cite{zheng2025easyr1} framework, Qwen2.5-7B~\cite{bai2025qwen2} as the base model, and Geometric3K~\cite{lu2021inter} as the dataset.
For code generation task, we use Code-R1~\cite{code-r1} framework, Qwen2.5-7B-Instruct-1M as the base model, and code-r1-12k\footnote{https://huggingface.co/datasets/ganler/code-r1-12k} as the dataset.
Both visual reasoning and code generation tasks are conducted on 8 GPU.

\paragraph{Hyper-parameters}

We mainly follow the default configuration of the \textit{verl} framework.
For sampling parameters in training generation, we set temperature to 1, top-p to 1 to encourage exploration, sequence length to 4096 for Qwen-series model and 8192 for DS-1.5B.
For sampling parameters in test generation, we set temperature to 0.7, top-p to 0.8, top-k to 20 and sequence length to 4096 for all models.
For optimization, training employs the Adam optimizer~\cite{kingma2014adam} with a constant learning rate and a linear warm-up over the first 10 steps.
For GRPO hyper-parameters, we set the clip ratio high to $0.28$ and clip ratio lower to $0.2$ following DAPO~\cite{yu2025dapo}.
All models are trained for 10 epochs.
The baseline method, 16-GRPO, is trained with batch sizes of 32 (32 prompts and 16 rollouts per prompt) and a learning rate $1\times 10^{-6}$.
As discussed in Appx.~\ref{appx:gradient_estimate},
we trained 2-GRPO with a larger batch size of 256 (256 prompts and 2 rollouts per prompt).
Both case will have 512 rollouts in each mini-batch of training. 
Since we have fewer update steps due to the larger batch size, 
we adjust the learning rate of 2-GRPO to $8\times 10^{-6}$
based on the linear relationship of learning rate and batch size~\cite{goyal2017accurate}.

\subsection{The Connection Between Training Rollouts and Computational Cost}
\label{appx:why_rollouts}

In Sec.~\ref{sec:main_exp}, the total number of rollouts generated and utilized during training is adopted as a metric for comparing the computational cost of different methods.

The rationale for this choice is as follows. A principled measure of computational cost in the context of RL post-training is the number of floating-point operations (FLOPs) performed. Unlike wall-clock time, which is susceptible to variations arising from software implementation details (e.g., optimization of training libraries) and hardware characteristics (e.g., GPU/CPU architecture, I/O throughput), FLOPs provide a more direct and stable measure of computational effort.

For a fixed base model and the same type of RL algorithm (GRPO in our case), 
the FLOPs required for a single forward or backward pass with one input prompt can be considered constant, for both the generation and training phases. 
Accordingly, the total number of rollouts executed during training is directly proportional to the FLOPs executed, 
thereby serving as a theoretically justified and consistent proxy for computational cost.

\subsection{Sample Discard Rate of 2-GRPO}
\label{appx:discard_rate}

As discussed in Sec.~\ref{sec:2-GRPO}, 2-GRPO may suffer from a high discard rate when prompts are either extremely easy or extremely difficult for the LLM.

In the RLVR setting, the average discard rate $P_{\text{discard}}$ can be estimated from the average reward $\bar{r}$ as
\[
P_{\text{discard}} = 1 - \bar{r}^2 - (1-\bar{r})^2.
\]

To quantify this effect, we report the average discard rate of 2-GRPO using Qwen-7B post-trained on MATH as a representative case (shown in Table~\ref{tab:discard_rate}).

\begin{table}[htbp]
\centering
\caption{The discard rate vs. training steps of 2-GRPO for Qwen-7B post-trained on MATH dataset.}
\vspace{1em}
\label{tab:discard_rate}
\begin{tabular}{cccccc}
\toprule
Step & 10 & 40 & 70 & 100 & 130 \\
\midrule
$P_\text{discard}$ & 0.5284 & 0.6917 & 0.6763 & 0.7711 & 0.7141 \\
\bottomrule
\end{tabular}
\end{table}

\subsection{Ablation Study on Group Size}
\label{appx:ablation_group_size}

\begin{table}[t!]
    \centering
    \caption{Ablation study on group size $G$: post-trained on MATH and DAPO, respectively, and evaluated on five mathematical reasoning benchmarks. M/P@32 stands for Mean@32 and Pass@32. \#P and \#R denote the number of prompts and the number of rollouts in a training mini-batch.}
    \label{tab:ablation}
    \vspace{0.5em}
    \setlength{\tabcolsep}{4.5pt}
    \renewcommand{\arraystretch}{1.15}
    \resizebox{\linewidth}{!}{%
    \begin{tabular}{@{} l l c c c c c c c @{}}
        \toprule
        \textbf{Model} & \textbf{$G$} & \textbf{Batch (\#P/\#R)}
            & \textbf{MATH-500} & \textbf{AMC 2023} & \textbf{Minerva Math}
            & \textbf{AIME 2025} & \textbf{Olympiad Bench} & \textbf{Average} \\
        \cmidrule(lr){4-8} \cmidrule(lr){9-9}
        & & & \multicolumn{6}{c}{\textit{Mean@32 / Pass@32}} \\
        \midrule

        \multicolumn{9}{@{}l}{\cellcolor{gray!15}\textit{\textbf{Post-training on MATH dataset}}} \\[2pt]

        \multicolumn{9}{@{}l}{\quad\textbf{Qwen-1.5B}} \\[1pt]
        \quad & w/o    & --
            & 31.83 / 81.92 & 34.30 / 79.23 & 5.33 / 28.91
            & 3.64 / 22.31  & 15.40 / 37.16 & \textbf{18.10 / 49.91} \\
        \quad & 2\,\textsuperscript{\textdagger} & 256/512
            & 69.28 / 87.43 & 49.53 / 81.76 & 16.25 / 33.26
            & 9.48 / 32.88  & 22.31 / 37.24 & \textbf{33.37 / 54.51} \\
        \quad & 2  & 32/64
            & 67.73 / 87.85 & 53.28 / 86.21 & 14.15 / 34.02
            & 6.15 / 29.54  & 23.11 / 37.82 & \textbf{32.88 / 55.09} \\
        \quad & 4  & 32/128
            & 69.05 / 87.49 & 52.50 / 92.01 & 15.29 / 33.57
            & 8.33 / 27.13  & 23.08 / 38.99 & \textbf{33.65 / 55.84} \\
        \quad & 8  & 32/256
            & 69.34 / 86.05 & 51.64 / 83.96 & 14.60 / 32.63
            & 7.18 / 32.24  & 22.77 / 36.69 & \textbf{33.11 / 54.31} \\
        \quad & 16 & 32/512
            & 70.24 / 87.24 & 51.25 / 83.46 & 16.84 / 33.46
            & 10.10 / 35.82 & 22.30 / 38.33 & \textbf{34.15 / 55.66} \\[3pt]

        \multicolumn{9}{@{}l}{\quad\textbf{Qwen-7B}} \\[1pt]
        \quad & w/o    & --
            & 47.16 / 85.95 & 38.36 / 85.29 & 5.99 / 31.10
            & 5.00 / 25.17  & 9.83 / 34.30  & \textbf{21.27 / 52.36} \\
        \quad & 2\,\textsuperscript{\textdagger} & 256/512
            & 75.23 / 89.77 & 64.60 / 81.53 & 23.13 / 38.45
            & 12.81 / 38.85 & 26.39 / 40.20 & \textbf{40.43 / 57.76} \\
        \quad & 2  & 32/64
            & 74.41 / 89.25 & 63.83 / 89.58 & 21.53 / 37.72
            & 11.67 / 33.05 & 26.04 / 41.34 & \textbf{39.50 / 58.19} \\
        \quad & 4  & 32/128
            & 76.24 / 88.16 & 63.51 / 84.97 & 23.09 / 41.03
            & 10.83 / 32.42 & 26.25 / 40.78 & \textbf{39.98 / 57.47} \\
        \quad & 8  & 32/256
            & 75.12 / 89.53 & 64.38 / 88.63 & 22.24 / 35.94
            & 12.71 / 35.85 & 26.25 / 40.52 & \textbf{40.14 / 58.09} \\
        \quad & 16 & 32/512
            & 75.90 / 88.24 & 61.79 / 80.77 & 22.81 / 37.68
            & 13.23 / 34.22 & 25.99 / 40.11 & \textbf{39.94 / 56.20} \\

        \midrule

        \multicolumn{9}{@{}l}{\cellcolor{gray!15}\textit{\textbf{Post-training on DAPO-Math-Sub dataset}}} \\[2pt]

        \multicolumn{9}{@{}l}{\quad\textbf{Qwen-1.5B}} \\[1pt]
        \quad & w/o    & --
            & 31.83 / 81.92 & 34.30 / 79.23 & 5.33 / 28.91
            & 3.64 / 22.31  & 15.40 / 37.16 & \textbf{18.10 / 49.91} \\
        \quad & 2\,\textsuperscript{\textdagger} & 256/512
            & 68.81 / 87.36 & 52.19 / 85.77 & 16.79 / 33.61
            & 8.13 / 29.33  & 23.52 / 39.29 & \textbf{33.89 / 55.07} \\
        \quad & 2  & 32/64
            & 67.71 / 87.68 & 53.82 / 88.35 & 16.85 / 34.83
            & 8.12 / 32.99  & 23.21 / 39.26 & \textbf{33.94 / 56.62} \\
        \quad & 4  & 32/128
            & 69.14 / 87.78 & 54.69 / 86.88 & 17.53 / 35.74
            & 8.43 / 36.18  & 23.30 / 39.00 & \textbf{34.62 / 57.12} \\
        \quad & 8  & 32/256
            & 70.25 / 86.84 & 57.57 / 81.19 & 17.80 / 35.08
            & 8.54 / 29.42  & 24.23 / 39.95 & \textbf{35.68 / 54.50} \\
        \quad & 16 & 32/256
            & 70.66 / 87.03 & 56.56 / 85.53 & 18.00 / 34.16
            & 9.58 / 32.31  & 24.55 / 39.19 & \textbf{35.87 / 55.64} \\[3pt]

        \multicolumn{9}{@{}l}{\quad\textbf{Qwen-7B}} \\[1pt]
        \quad & w/o    & --
            & 47.16 / 85.95 & 38.36 / 85.29 & 5.99 / 31.10
            & 5.00 / 25.17  & 9.83 / 34.30  & \textbf{21.27 / 52.36} \\
        \quad & 2\,\textsuperscript{\textdagger} & 256/512
            & 77.43 / 90.51 & 64.84 / 91.59 & 21.95 / 38.05
            & 14.58 / 33.03 & 29.86 / 45.24 & \textbf{41.73 / 59.68} \\
        \quad & 2  & 32/64
            & 75.24 / 89.37 & 66.33 / 90.56 & 23.49 / 39.89
            & 15.21 / 41.71 & 28.60 / 43.04 & \textbf{41.77 / 60.91} \\
        \quad & 4  & 32/128
            & 76.58 / 90.62 & 66.33 / 95.75 & 23.44 / 40.06
            & 15.21 / 40.02 & 27.94 / 42.24 & \textbf{41.90 / 61.74} \\
        \quad & 8  & 32/256
            & 72.43 / 87.27 & 71.17 / 93.08 & 25.03 / 37.94
            & 17.71 / 39.67 & 28.74 / 41.46 & \textbf{43.02 / 59.88} \\
        \quad & 16 & 32/512
            & 77.35 / 88.79 & 69.69 / 87.31 & 24.45 / 40.04
            & 14.27 / 33.73 & 28.86 / 39.84 & \textbf{42.92 / 57.94} \\

        \bottomrule
        \multicolumn{9}{@{}l}{%
            \textsuperscript{\textdagger}2-GRPO with larger batch size (256 prompts / 512 rollouts).
        }
    \end{tabular}}
\end{table}

\begin{figure*}[h!]
    \centering
    \begin{subfigure}{0.49\linewidth}
        \centering
        \includegraphics[width=\linewidth]{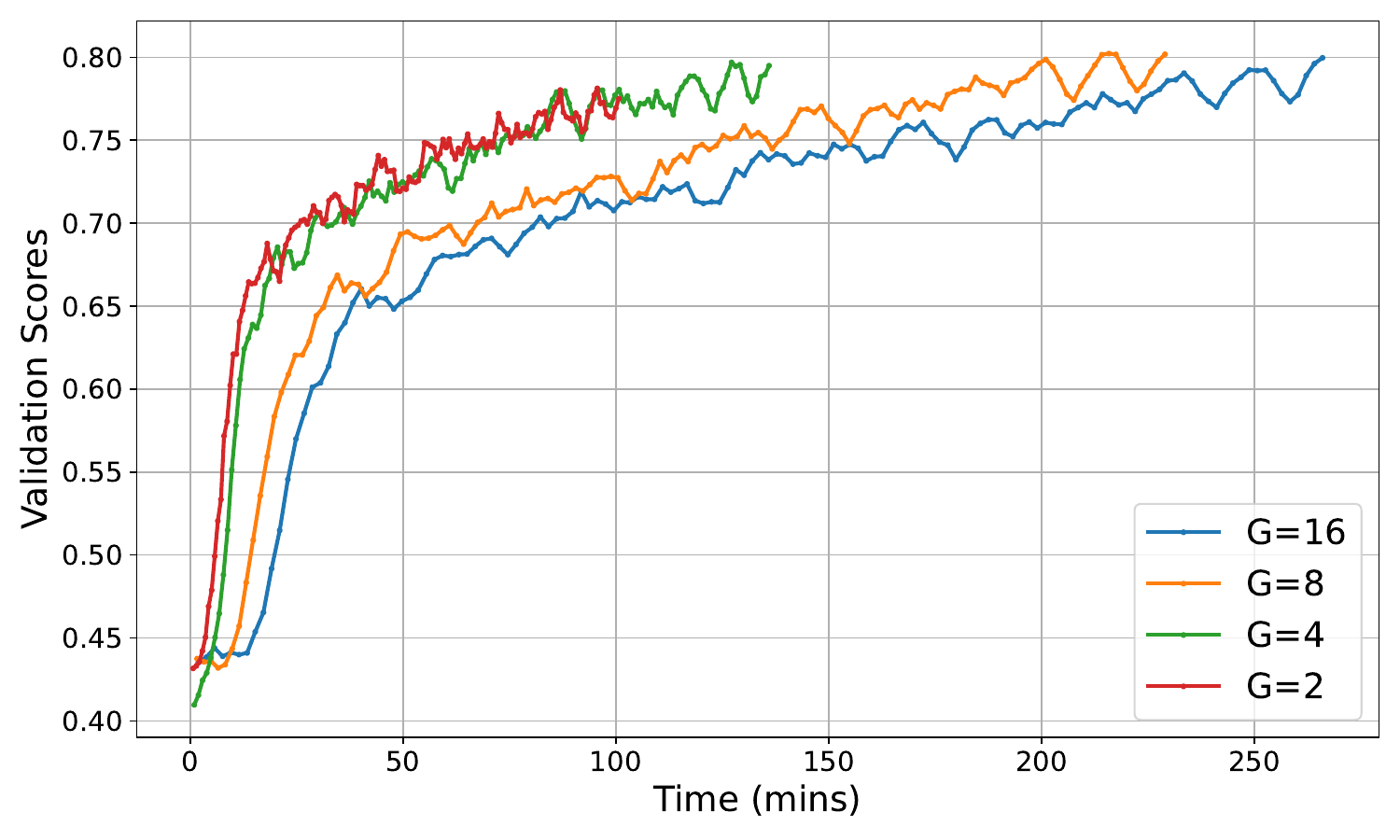}
        \caption{Training reward during training.}
        \label{fig:ctr_train_reward}
    \end{subfigure}\hfill
    \begin{subfigure}{0.49\linewidth}
        \centering
        \includegraphics[width=\linewidth]{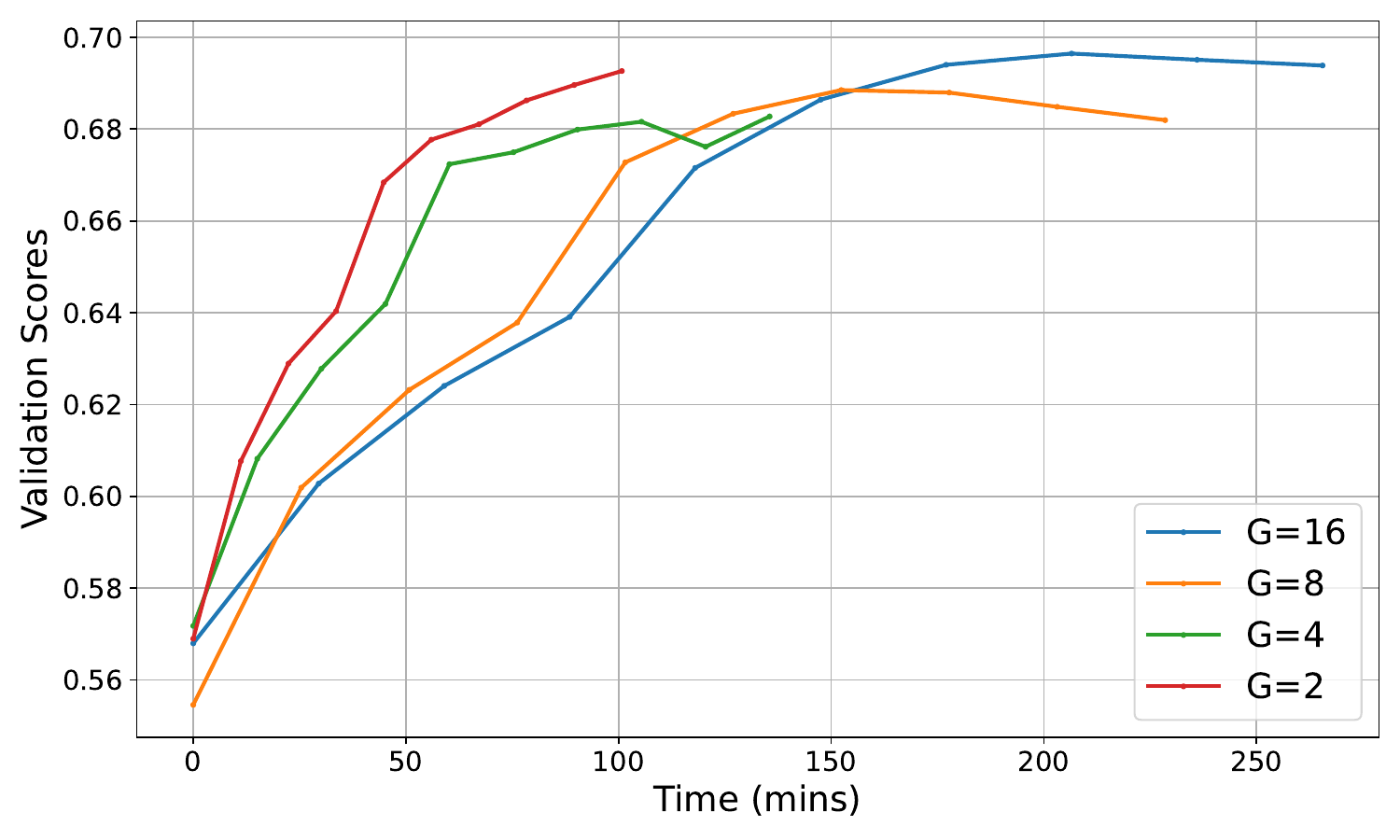}
        \caption{Evaluation score on the Test set.}
        \label{fig:ctr_test_score}
    \end{subfigure}
    \caption{Reward curves and validation scores of different group size on MATH. Curves are post simple-moving-average (SMV) with window-size=4 for better visualization, respectively.}
    \label{fig:dynamcis_group_size}
\end{figure*}
We present a comprehensive ablation study on group size in Table~\ref{tab:ablation}.
To further illustrate this effect, Figure~\ref{fig:dynamcis_group_size} reports the reward curves during training alongside the corresponding validation scores throughout the training process on MATH dataset.

\section{Limitation}
The contrastive learning nature of GRPO applies regardless of whether rewards are continuous or binary. 
However, the present study focuses primarily on the reasoning tasks with the RLVR setting, 
and we leave the empirical investigation of continuous rewards to future work due to the space limit.

\section{The Use of Large Language Models (LLMs)}
We used LLMs in writing, editing and formatting purposes.
Our experiments also involve the LLMs.







\end{document}